\documentclass[11pt]{article}

\usepackage{amsmath}
\usepackage{amssymb}
\usepackage{float}
\usepackage{graphicx}

\usepackage[comma,sort&compress]{natbib}
\usepackage{color} 

\usepackage{lineno}
\usepackage{url}
\usepackage{pdfpages}
\usepackage{setspace} 
\usepackage{lineno}

\usepackage[table,x11names]{xcolor}
\usepackage{blkarray}
\usepackage{amsmath,amstext,amsthm,amssymb}
\usepackage{authblk}

\usepackage{hyperref}

\usepackage[table]{xcolor}

\hypersetup{
	bookmarks=true,         
	unicode=false,          
	pdftoolbar=true,        
	pdfmenubar=true,        
	pdffitwindow=false,     
	pdfstartview={FitH},    
	pdftitle={My title},    
	pdfauthor={Author},     
	pdfsubject={Subject},   
	pdfcreator={Creator},   
	pdfproducer={Producer}, 
	pdfkeywords={keywords}, 
	pdfnewwindow=true,      
	colorlinks=true,       
	linkcolor=blue,          
	citecolor=blue,        
	filecolor=magenta,      
	urlcolor=cyan           
}

\usepackage[labelfont=bf,labelsep=period,justification=raggedright]{caption}

\makeatletter
\renewcommand{\@biblabel}[1]{\quad#1.}
\makeatother

\date{}

\title{\bf{Co-evolution of social reward and punishment under institutional interventions
} }

\author[1,2]{Van An Nguyen}
\author[1,2]{Vuong Khang Huynh}
\author[1,2]{Hoai Thuong Nguyen}
\author[1,2]{Duc Tin Duong}

\author[1,2]{An Nguyen Gia}
\author[1,2]{Tat Kien Nguyen}
\author[1,2]{Huu Loi Bui}
\author[1,2]{My Nguyen Tra}
\author[1,2]{Ho Nam Duong}
\author[1,2]{Ba Thanh Phan}
\author[1,2]{Thanh Vo}
\author[1,2]{Dinh Anh Trung Hoang}
\author[3]{Adeela Bashir}
\author[3]{Zhao Song}

\author[4]{Manh Hong Duong} 
\author[1,2,$\star$]{Le Hong Trang}
\author[3,$\star$]{The Anh Han }

\affil[1]{Faculty of Computer Science and Engineering, Ho Chi Minh City University of
Technology (HCMUT), Vietnam}
\affil[2]{Vietnam National University - Ho Chi Minh City (VNU-HCM), Vietnam}
\affil[3]{School of Computing, Engineering and Digital Technologies, Teesside University, United Kingdom}
\affil[4]{School of Mathematics, University of Birmingham, Birmingham, United Kingdom}
\affil[$\star$]{Corresponding authors: The Anh Han (Email: t.han@tees.ac.uk), Le Hong Trang (Email: lhtrang@hcmut.edu.vn)}

\begin{document}
	\maketitle
  

\section*{Abstract}

We investigate how peer and institutional incentives jointly shape the evolution of cooperation, social welfare, and enforcement efficiency in social dilemmas. In a Prisoner’s Dilemma with four strategies, unconditional cooperators (C), defectors (D), social punishers (SP), and social rewarders (SR), we allow decentralised peer incentives and centralised institutional incentives to act simultaneously, with the institution able to reward or punish any subset of strategies. In infinite well-mixed populations, we analyse the resulting four-strategy replicator dynamics, and in structured populations we use agent-based simulations on square lattices to study spatial effects and network reciprocity. Intervention schemes are evaluated by equilibrium states and evolutionary flow for infinite well-mixed populations, by cooperation levels and social welfare for structured populations, defined as aggregate population payoff net of institutional cost. We find that peer punishment most strongly promotes cooperation, whereas peer reward is more beneficial for social welfare. Institutionally rewarding peer incentive strategies substantially improves both cooperation and welfare, while subsidising unconditional cooperators has little impact. Under institutional punishment, directly penalising defectors is the only consistently effective policy; punishing peer incentive strategies dismantles decentralised incentives, reduces cooperation, and harms social welfare, showing that maximising cooperation does not necessarily optimise overall societal benefit. Our findings provide design principles for institutions seeking to balance cooperation promotion with welfare maximisation. \\

 \noindent \textbf{Keywords:} Evolution of cooperation, social dilemma, peer incentives, social welfare, reward, punishment,   evolutionary dynamics.

 \tableofcontents

\section{Introduction}

The evolution and maintenance of cooperation among self-regarding individuals is a longstanding puzzle across the behavioural, social, and computational sciences \citep{nowak2006,perc2017statistical,sigmund2010calculus}. This tension is classically captured by the Prisoner's Dilemma (PD), where mutual cooperation yields the highest collective payoff, yet each individual is tempted to defect and free-ride on the contribution of others. A large body of work in evolutionary game theory has identified mechanisms that can tip the balance toward cooperation, including kin and group selection, direct and indirect reciprocity, spatial structure, and reward and punishment \citep{boyd2003evolution,fehr2002altruistic,santos2006evolutionary, Sigmund2001PNAS,nowak2005evolution,hamilton1964genetical}.

Incentive mechanisms have been widely studied as a way to promote cooperation in social dilemmas. By modifying the payoff structure of interactions, these mechanisms can shift the balance between cooperative and selfish behaviours, increasing the evolutionary advantage of cooperation and enabling its persistence in populations \citep{han2018cost, sigmund2010calculus, wang2019exploring,Sigmund2001PNAS}. Both forms of intervention are pervasive in real societies, from informal social sanctioning among peers to formal legal and regulatory enforcement, and understanding their interplay is key to designing effective institutions for collective action \citep{bowles2002social, ostrom1990governing,van2014reward}.

These existing evolutionary models of incentivised cooperation focus on a single type of incentive at a time, i.e. either social punishment or social reward—coexisting alongside unconditional cooperators and defectors \citep{perc2017statistical,HAN2026_social_welfare}. Thus, an important question is how punishment and reward co-evolve when both are simultaneously available in the same population. In real societies, these peer incentive strategic choices can coexist, compete, mutually reinforce one another, or drive each other to extinction. Moreover, although institutional incentives have been widely studied in simple two-strategy settings (cooperators and defectors), their impact when layered onto a richer strategy space that already includes peer-based incentives, where punishers and rewarders themselves may become targets of institutional intervention, remains largely unexplored.

\begin{figure}[H]
    \centering
    \includegraphics[width=\linewidth]{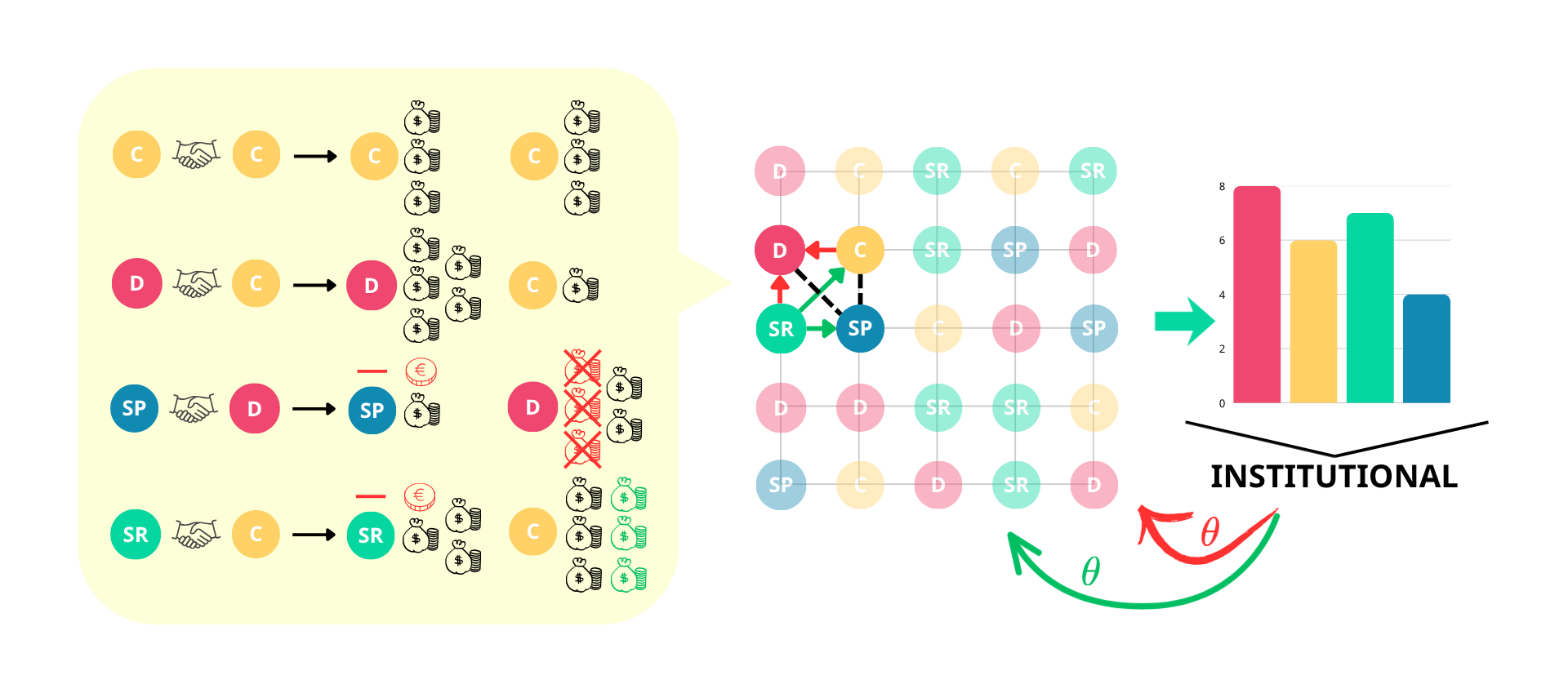}
    \caption{\textbf{Overall illustration of overall approach in this work}. (Left) The four strategies and their pairwise interaction. Two cooperators ($C$) each obtain the mutual-cooperation payoff; a defector ($D$) exploits a cooperator, gaining a higher payoff while $C$ receives the sucker's payoff. A social punisher ($SP$) pays a cost $\epsilon$ to reduce a defector's payoff by $\delta$. (Centre) Players occupy the nodes of a two-dimensional lattice with periodic boundary conditions and interact only with their four nearest neighbours; arrows illustrate a focal interaction, with punishment (red) directed to $D$ and reward (green) directed to $C$. $SR$ plays an important role in increasing the social welfare but being prone to extinction with highly $\epsilon$. (Right) An external institution observes the population composition and allocates a budget $\theta$ to reward or punish to a chosen set of cooperative as well as defective target types.}
    \label{fig:simulation_overview}
\end{figure}

In this paper, we address this crucial gap by extending the classical three-strategy peer-incentive model of the social dilemmas \citep{szolnoki2010reward,han2016emergence,Sigmund2001PNAS} to a four-strategy model comprising unconditional cooperators ($C$), unconditional defectors ($D$), social punishers ($SP$), and social rewarders ($SR$), who coexist and compete within the same population (see Figure \ref{fig:simulation_overview}). Building on this enriched strategy space, we further incorporate institutional-level incentives, allowing an external institution to reward or punish any combination of strategies, including the peer-incentivising types themselves. This reflects the idea that peer punishment and peer reward may, in some contexts, be viewed by institutions as undesirable behaviours (e.g., vigilantism or bribery) worth discouraging.

We study the resulting evolutionary dynamics using the framework of evolutionary game theory (EGT) \citep{sigmund2010calculus}. Specifically, we model an infinite, well-mixed population and describe the evolution of strategy frequencies using the replicator dynamics. We characterise the equilibria of this dynamical system, including the monomorphic (single-strategy) fixed points as well as boundary and interior equilibria involving the coexistence of multiple strategies, and we determine their stability, in particular asymptotically stable, unstable, or neutrally stable, via linear stability analysis of the corresponding Jacobian, complemented where necessary by center-manifold analysis at non-hyperbolic points. This analysis provides a systematic characterisation of the long-run outcomes of the four-strategy peer-and-institutional-incentive model and clarifies the conditions under which cooperation, defection, or a mixture of incentivising strategies prevail in the infinite-population limit.

This work further investigates these strategies through structured population simulations, where individuals interact within a finite network environment \citep{szabo2007evolutionary}. This allows us to observe spatial effects and dynamic behaviours that may not be captures by theoretical analysis alone.

\section{Literature review}
The evolution of cooperation has long been a central topic in evolutionary game theory, where social dilemmas such as the Prisoner's Dilemma and Public Goods Game capture the conflict between individual rationality and collective welfare. In one-shot interactions, defection is the dominant strategy despite mutual cooperation producing higher collective benefits. This apparent contradiction has motivated extensive research into mechanisms that enable cooperation to emerge and persist, including kin selection, reciprocity, spatial structure, and institutional regulation \citep{nowak2006,sigmund2010calculus,perc2017statistical}.

Among these mechanisms, peer incentives have received considerable attention. Under peer punishment, cooperators incur a personal cost to reduce the payoff of defectors, whereas under peer reward they pay a cost to increase the payoff of other cooperators \citep{wang2013impact}. Experimental and theoretical studies have shown that peer punishment can effectively sustain cooperation even among unrelated individuals \citep{fehr2002altruistic}, while peer reward can achieve comparable or even better outcomes with lower social cost under certain conditions \citep{rand2009positive,Sigmund2001PNAS}. Since both mechanisms require individuals to bear additional costs, their effectiveness has generally been evaluated by their ability to promote cooperation, although the resulting impact on the population's overall payoff is not always positive.

Beyond decentralized peer interactions, another major line of research has investigated institutional incentives, where an external authority expends resources to influence individual behaviour. Early studies examined how institutional rewards, punishments, or hybrid incentive schemes affect the stability of cooperative behaviour \citep{szolnoki2010reward}. More recent work introduced adaptive institutional policies in which external investment depends on the current state of the population, allowing analytical derivation of strategies that achieve a desired cooperation level while minimizing intervention cost \citep{duong2021cost,duong2023cost,duong2024cost}.

Institutional incentive models have subsequently been extended to structured populations, where interactions occur primarily among neighbouring individuals rather than uniformly across the population. In such settings, institutions must determine not only the amount of investment but also where incentives should be allocated, since different regions of the network may exhibit different levels of cooperation. Previous studies have shown that localized and neighbourhood-dependent interventions often outperform globally uniform strategies by allocating resources more efficiently and achieving higher levels of cooperation \citep{han2018fostering,wang2022decentralized,wang2023optimization}.

As institutional models have become increasingly sophisticated, researchers have begun to consider evaluation criteria beyond cooperation frequency alone. While many studies measure the effectiveness of interventions by the cooperation level achieved or by the external cost required to sustain cooperation, these metrics do not necessarily reflect the overall benefit to the population. Recent work has argued that social welfare, defined as the aggregate population payoff after accounting for the cost of institutional intervention, provides a more comprehensive evaluation criterion because it simultaneously captures the benefits of cooperation and the resources required to maintain it \citep{han2025cooperation}. This perspective highlights that an intervention producing higher cooperation is not necessarily preferable if the associated enforcement cost outweighs its benefits.

Evolutionary analyses of these incentive mechanisms are commonly conducted using replicator dynamics, which describe how strategy frequencies evolve according to their payoff advantage relative to the population average \citep{russell2026convergence}. Existing analytical studies have characterised the equilibrium structure and stability of populations containing unconditional cooperators, defectors, and peer-based incentive strategies, identifying the conditions under which cooperation can emerge and persist \citep{sasaki2015voluntary}. Complementary agent-based simulations have further demonstrated how network structure influences the effectiveness of institutional interventions and the resulting evolutionary outcomes \citep{han2018fostering,wang2022decentralized,wang2023optimization}.

Despite these significant advances, most existing studies continue to evaluate institutional incentive schemes primarily from the perspectives of maximizing cooperation or minimizing intervention cost. Although these objectives are important, they do not necessarily identify the intervention that maximizes the overall benefit to society. Different stable equilibria may exhibit similar cooperation levels while producing substantially different net social welfare once the costs of maintaining cooperation are taken into account \citep{han2025cooperation}. Consequently, a systematic comparison of institutional intervention strategies under a unified social welfare framework remains limited, particularly across both analytical well-mixed models and simulation-based structured populations. Addressing this gap is essential for understanding whether strategies designed to maximize cooperation also maximize overall social welfare.
\section{Model and Methods}

In this section, we recall the Prisoner’s Dilemma and the definition of the peer and institutional incentives, and then formulate the corresponding payoff matrices, replicator dynamics for well-mixed populations, and agent-based simulation framework on square lattices used in our analysis.

\subsection{Prisoner's Dilemma }
Players interact with each other via the one-shot Prisoner's Dilemma (PD) game, players choose their strategies - either to cooperate ($C$) or to defect ($D$) - to play with each other, with payoffs given by the following matrix:
\begin{equation}
    \begin{blockarray}{ccc}
    & C & D\\
    \begin{block}{c(cc)}
      C & R & S \\
      D & T & P \\
    \end{block}
  \end{blockarray}.
\end{equation}

If both interacting players follow the same strategy, they receive the same payoff: reward $R$ for mutual cooperation and punishment $P$ for mutual defection. If the agents play different strategies, the cooperator gets the sucker's payoff $S$, and the defector gets the temptation to defect $T$. In a PD situation, the parameters of the matrix must satisfy the ordering $T>R>P>S$ \citep{coombs1973reparameterization}. 

 {The strength of the dilemma in the PD game can be varied adopting a simplified scaling approach from  \citep{wang2015universal,ito2018scaling,arefin2020social}. Indeed, by fixing $T - R =P - S = 1$, the dilemma strength decreases when  $R - P$ increases. }

\subsection{Peer incentives}
\label{sect:peer_incentive}
Peer incentives refer to decentralized mechanisms where individuals directly influence the payoff of others through costly rewarding or punishment actions. In the Prisoner's Dilemma game, social punishers ($SP$) and social rewarders ($SR$) cooperate with others while influencing their payoffs through peer-based incentives. Specifically, they decrease or increase the payoff of other players by an amount of $\delta_P$ (resp., $\delta_R$) at a personal cost of $\epsilon_P$ (resp., $\epsilon_R$). 

We treat the peer cost $\epsilon$ and impact $\delta$ as independent parameters rather than reporting only their ratio, since they govern distinct channels: $\epsilon$ alone determines the enforcers' fitness disadvantage and hence the dynamics, whereas $\delta$ enters welfare directly. Peer reward contributes positively when $\delta > \epsilon$, while peer punishment destroys value twice over - the punisher pays and the punisher loses, with no one receiving - and so contributes negatively throughout. Results are therefore reported on the $(\epsilon, \delta)$ grid for each target set and each institutional impact.

We consider a minimal model of peer incentives in a one-shot Prisoner's Dilemma game, where four strategies are available: unconditional cooperators ($C$), unconditional defectors ($D$), social punishers ($SP$), and social rewarders ($SR$). The payoff matrix describing the interactions among these four strategies is presented below. 


\begin{equation}
 P =
\begin{blockarray}{ccccc}
  & C & D & SP & SR\\
    \begin{block}{c(cccc)}
      C & R & S & R & R + \delta_R \\
      D & T & P & T -\delta_P & T \\
       SP & R & S -\epsilon_P & R & R + \delta_R  \\
       SR & R - \epsilon_R & S & R - \epsilon_R & R - \epsilon_R + \delta_R \\
    \end{block}
  \end{blockarray} \ \ \ \
  \label{eq:payoff_matrix_without_institution}
\end{equation}

\subsection{Institutional incentives}
Institutional incentives are mechanisms in which an external third party - an institution - expends a budget to alter players' payoffs so as to promote cooperation. Unlike peer incentives, the institution does not participate in the game and is not subject to selection, it therefore always acts and cannot be outcompeted. In general, the institution pays $|\theta_i|$ per targeted individual to shift strategy $i$'s payoff by $\theta_i$: $\theta_i>0$ is a net reward, $\theta_i<0$ a net punishment, and each of the four strategies can be targeted independently ($\theta_C,\theta_D,\theta_{SP},\theta_{SR}\in\mathbb R$). This is the fully general model used for the equilibrium analysis below.

For the simulation comparisons in the Results section, we specialise this general model to institutional \emph{reward} restricted to the cooperative types: $\theta_D=0$ always, and only strategies in a chosen target set $\mathcal T \subseteq \{C,SP,SR\}$ receive $\theta_i>0$ (all strategies outside $\mathcal T$, including $D$, get $\theta_i=0$). All three of $C$, $SP$ and $SR$ cooperate, yet differ in post-game behaviour and welfare contribution. We therefore examine all seven non-empty target sets: the single-target policies $\{C\}$, $\{SP\}$, $\{SR\}$; the dual-target policies $\{C,SP\}$, $\{C,SR\}$, $\{SP,SR\}$; and the full-target policy $\{C,SP,SR\}$. In terms of Monte Carlo step, reward and punishment are simply creating the difference enforcing players to change to the biased policies, but in term of social welfare, this can introduce new insights. In order to do so, we setup institutional \emph{punishment} targetting non pure-cooperators: $\theta_C = 0$ as the control scenario. We observe default no punishment added to make sure both version of institutional incentives operate, then we consider seven target sets: the single-target policies $\{D\}$, $\{SP\}$, $\{SR\}$; the dual-target policies $\{D,SP\}$, $\{D,SR\}$, $\{SP,SR\}$; and the full-target policy $\{D,SP,SR\}$. By doing so, we will test the model against similar enforcing scenario in term of updating mechanism, but differ in final welfare obtained.

\subsubsection*{Payoff matrix with institutional incentives}
We model institutional reward/punishment as a fixed payoff shift applied uniformly to every interaction of a given strategy: the institution adds $\theta_C$ to every payoff earned by a $C$-player, $\theta_D$ to every payoff earned by a $D$-player, and likewise $\theta_{SP}$, $\theta_{SR}$ for $SP$ and $SR$ (a positive $\theta_i$ is a net reward, a negative $\theta_i$ a net punishment, targeting strategy $i$ regardless of its opponent). In matrix terms this means adding the constant $\theta_i$ to every entry of row $i$ of the peer-incentive payoff matrix $P$ from Section \ref{sect:peer_incentive}
\begin{equation}
P^{\theta} = P + \theta\,\mathbf{1}^\top, \qquad \theta = (\theta_C,\theta_D,\theta_{SP},\theta_{SR})^\top, \quad \mathbf{1} = (1,1,1,1)^\top,
\end{equation}
i.e.\ explicitly (rows/columns ordered $C,D,SP,SR$)
\begin{equation}
P^{\theta} =
\begin{pmatrix}
R+\theta_C & S+\theta_C & R+\theta_C & R+\delta_R+\theta_C \\
T+\theta_D & P+\theta_D & T-\delta_P+\theta_D & T+\theta_D \\
R+\theta_{SP} & S-\epsilon_P+\theta_{SP} & R+\theta_{SP} & R+\delta_R+\theta_{SP} \\
R-\epsilon_R+\theta_{SR} & S+\theta_{SR} & R-\epsilon_R+\theta_{SR} & R-\epsilon_R+\delta_R+\theta_{SR}
\end{pmatrix}.
\end{equation}

\subsection{Replicator dynamics for the prisoner's dilemma with peer incentives}
Let $f=[x,y,z,w]^\top$ denote the population state vector, where $x$, $y$, $z$, and $w$ represent the frequencies of strategies $C$, $D$, $SP$, and $SR$, respectively, satisfying the simplex constraint $\mathbf{1}^\top f = 1$. Under the assumption of random matching, the expected payoff of each strategy is determined entirely by the current population composition and the underlying payoff matrix.

We begin by considering the baseline game without institutional incentives. The expected payoff associated with each strategy can be expressed as follows:

\begin{equation}
\begin{pmatrix}\Pi_C \\ \Pi_D \\ \Pi_{SP} \\ \Pi_{SR}\end{pmatrix} = Pf,
\end{equation}
i.e.
\begin{align}
\Pi_C &= R + (S-R)\,y + \delta_R w \\
\Pi_D &= T + (P-T)\,y - \delta_P z \\
\Pi_{SP} &= \Pi_C - \epsilon_P y \\
\Pi_{SR} &= \Pi_C - \epsilon_R (1-y)
\end{align}
The average payoff of the population is given by
\begin{equation}
\bar{\Pi} \;=\; f^\top Pf \;=\; x\Pi_C + y\Pi_D + z\Pi_{SP} + w\Pi_{SR}
\;=\; (1-y)\,\Pi_C \;+\; y\,\Pi_D \;-\; \epsilon_P y z \;-\; \epsilon_R w(1-y)
\end{equation}

To characterize the evolutionary dynamics, we adopt the replicator equations, whereby the frequency of each strategy evolves according to its payoff advantage over the population average. Throughout, $x,y,z,w$ (and the vector $f=(x,y,z,w)$) are understood as functions of time $t$, and a dot denotes the time derivative, $\dot{(\cdot)} := d(\cdot)/dt$. Exploiting the simplex constraint to eliminate $x$, the four-dimensional system can be reduced to the following three-dimensional dynamical system:
\begin{align}
\dot{y} &= y\left(\Pi_D - \bar{\Pi}\right), \\
\dot{z} &= z\left(\Pi_{SP} - \bar{\Pi}\right), \\
\dot{w} &= w\left(\Pi_{SR} - \bar{\Pi}\right).
\end{align}
For notational convenience, we define the auxiliary payoff-difference function
\begin{equation}
F(y,z,w) \;=\; \Pi_D - \Pi_C \;=\; (T-R) + (R+P-T-S)\,y - \delta_P z - \delta_R w.
\end{equation}
Substituting this expression into the replicator equations yields the following explicit polynomial dynamical system
\begin{align}
\dot{y} &= y(1-y)\,F \;+\; \epsilon_P y^2 z \;+\; \epsilon_R\, y w (1-y), \\
\dot{z} &= -yz\,F \;+\; \epsilon_P y z (z-1) \;+\; \epsilon_R\, z w (1-y), \\
\dot{w} &= -yw\,F \;+\; \epsilon_P y z w \;-\; \epsilon_R\, w (1-y)(1-w).
\label{baseline-replicator}
\end{align}
\subsection{Replicator dynamics for the prisoner's dilemma with both peer and institutional incentives}
As the institutional incentive policies are implemented as a fixed and uniform shift to the payoff matrix, the strategy payoffs are simply the peer-incentive payoffs shifted by the corresponding $\theta_i$:
\begin{equation}
\begin{pmatrix}\Pi_C^\theta \\ \Pi_D^\theta \\ \Pi_{SP}^\theta \\ \Pi_{SR}^\theta\end{pmatrix} = P^\theta f = Pf + \theta,
\end{equation}
i.e.
\begin{align}
\Pi_C^\theta &= \Pi_C + \theta_C = R + (S-R)y + \delta_R w + \theta_C, \\
\Pi_D^\theta &= \Pi_D + \theta_D = T + (P-T)y - \delta_P z + \theta_D, \\
\Pi_{SP}^\theta &= \Pi_{SP} + \theta_{SP} = \Pi_C^\theta - \epsilon_P y + \alpha, \\
\Pi_{SR}^\theta &= \Pi_{SR} + \theta_{SR} = \Pi_C^\theta - \epsilon_R (1-y) + \beta,
\end{align}
where we introduce the incentive shifts relative to $C$
\begin{equation}
\alpha := \theta_{SP} - \theta_C, \qquad \beta := \theta_{SR} - \theta_C, \qquad \gamma := \theta_D - \theta_C,
\end{equation}
it follows that $\Pi_D^\theta = \Pi_C^\theta + F + \gamma$, with $F=F(y,z,w)$ as defined in Section \ref{sect:peer_incentive}.

The average population payoff is given as:
\begin{equation}
\bar\Pi^\theta = f^\top P^\theta f = f^\top P f + \theta^\top f (\mathbf 1^\top f) = \bar\Pi + \theta^\top f = \bar\Pi + x\theta_C + y\theta_D + z\theta_{SP} + w\theta_{SR}.
\end{equation}
Equivalently, in the $C$-relative shifts,
\begin{equation}
\Pi_C^\theta - \bar\Pi^\theta = -yF - y\gamma + \epsilon_P yz - z\alpha + \epsilon_R w(1-y) - w\beta.
\end{equation}

Consequently, the replicator equations is given as:
\begin{align}
\dot{x} &= x\left[-yF - y\gamma + \epsilon_P yz - z\alpha + \epsilon_R w(1-y) - w\beta\right] \\
\dot{y} &= y(1-y)(F+\gamma) + \epsilon_P y^2 z - yz\alpha + \epsilon_R yw(1-y) - yw\beta \\
\dot{z} &= -yz(F+\gamma) + \epsilon_P yz(z-1) + z\alpha(1-z) + \epsilon_R zw(1-y) - zw\beta \\
\dot{w} &= -yw(F+\gamma) + \epsilon_P yzw - zw\alpha - \epsilon_R w(1-y)(1-w) + w\beta(1-w)
\end{align}
Setting $\theta_C=\theta_D=\theta_{SP}=\theta_{SR}=0$ (so $\alpha=\beta=\gamma=0$) recovers exactly the purely peer-incentive dynamics. 

Note that we can perform the replicator dynamics in vector form
\begin{equation}
\label{eq:vector-form}
\dot f = \mathrm{diag}(f)\left(P^\theta f - (f^\top P^\theta f)\,\mathbf 1\right).
\end{equation}

\subsection{Agent-based simulations on network}

For agent-based simulation, we arrange players on a two-dimensional square lattice with periodic boundary conditions. One player sits on each node and plays only against the four neighbours in its von Neumann neighbourhood. We use a square lattice with $L = 100$ (resulting in an overall network size $Z = 100 \times 100 = 10000$). At the beginning of the simulation, we assign each player one of 4 strategies: $C$, $D$, $SR$, or $SP$, with equal probability. Each player then interacts with its neighbours to obtain the accumulated payoffs with values described in Equation \ref{eq:payoff_matrix_without_institution}. We set the specific values in the payoff matrix $R=3, T=5, S=0,$ and $P=1$ and run with $\epsilon_R=\epsilon_P = \epsilon, \delta_R = \delta_P = \delta$ ranging from $[0, 3], [0,3]$ respectively. We follow the strategy update process in \citep{SONG2026112303,bashir2026network} without mutation. We use the same imitation probability Fermi function, which gives the probability for agent A with payoff $f_A$ to imitate agent B with payoff $f_B$:
    \[
        P(A \leftarrow B) = \bigl(1 + e^{(f_A - f_B)/K}\bigr)^{-1},
    \]
where $K$ controls the noise level \citep{traulsen2006stochastic}. As a typical choice, we use $K=0.3$ for all experiments unless otherwise specified \citep{szabo2007evolutionary,perc2017statistical}. We run the simulation for $10 \times 10^4$ steps, with results averaged over the last $1000$ steps. 

In term of institutional interference, we run simulation for both incentive and punishment scenario. We use efficiency coefficient $a=1$ and per capita cost value $\theta = 1.0$ for all experiments unless further description.

\section{Results}
In this section, we present the findings of this paper. For well-mixed populations, we analyze the equilibrium properties of the two replicator dynamics and compare them with the simulation results. For square-lattice populations, we present numerical results based on agent-based simulations. 
\subsection{Well-mixed populations}

\begin{figure}[t]
    \centering
    \includegraphics[width=0.7\linewidth]{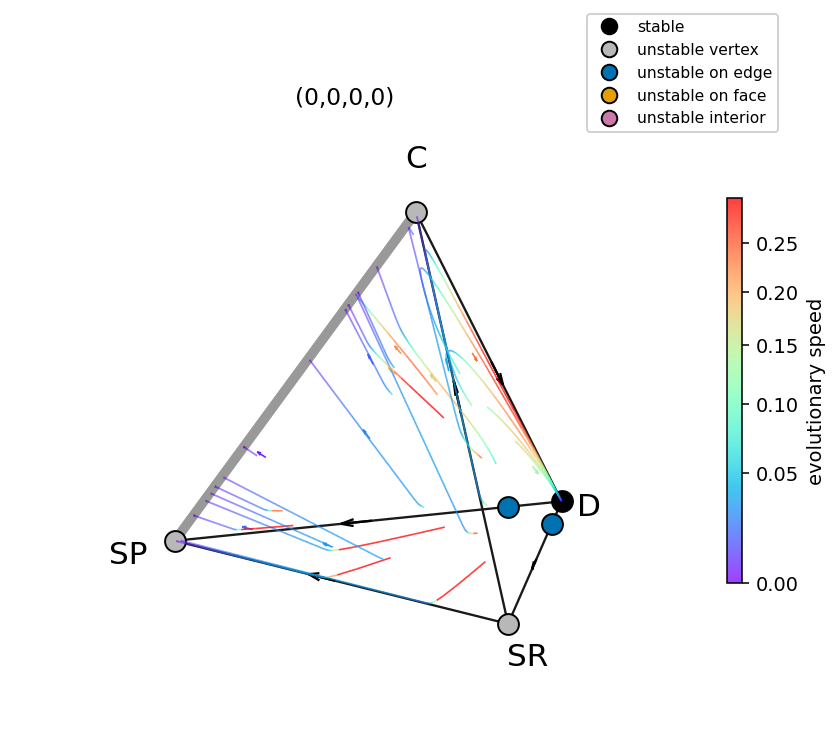}
    \caption{\textbf{Overview of the four strategies under pure social (peer) incentives.} The four strategies are unconditional cooperation ($C$), unconditional defection ($D$), social (peer) punishment ($SP$), and social (peer) reward ($SR$). This figure illustrates the case without institutional incentives, $(0,0,0,0)$, corresponding to $\theta_C = 0$, $\theta_D = 0$, $\theta_{SP} = 0$, and $\theta_{SR} = 0$.}
    \label{fig:well_mixed_overview}
\end{figure}
\subsubsection{Analytical Results - Peer Incentives Only}
This section investigates the evolutionary dynamics of a well-mixed infinite population consisting of four competing strategies: Cooperators ($C$), Defectors ($D$), Social Punishers ($SP$), and Social Rewarders ($SR$). 
The equilibrium points of the system are obtained by solving the system of equations in (\ref{baseline-replicator}). For polymorphic equilibria, excluding the four vertices of the simplex, the equilibrium conditions reduce to the linear system $\Pi_s=\bar{\Pi}$ for $s\in{C,D,SP,SR}$, indicating that some or all coexisting strategies must receive the same expected payoff. Section \ref{sect:insti_result} provides precise formulations to the solutions of the linear system.

Let $B=R+P-T-S$, such that $\partial F/\partial y=B$, $\partial F/\partial z=-\delta_P$, and $\partial F/\partial w=-\delta_R$. To assess the local stability of each equilibrium, we perform a linear stability analysis by evaluating the Jacobian matrix of the reduced dynamical system with respect to the variables $(y,z,w)$:

\begin{equation}
J =
\begin{pmatrix}
\dfrac{\partial \dot y}{\partial y} & \dfrac{\partial \dot y}{\partial z} & \dfrac{\partial \dot y}{\partial w} \\[8pt]
\dfrac{\partial \dot z}{\partial y} & \dfrac{\partial \dot z}{\partial z} & \dfrac{\partial \dot z}{\partial w} \\[8pt]
\dfrac{\partial \dot w}{\partial y} & \dfrac{\partial \dot w}{\partial z} & \dfrac{\partial \dot w}{\partial w}
\end{pmatrix}
\end{equation}

with entries

\begin{align}
\frac{\partial \dot y}{\partial y} &= (1-2y)F + y(1-y)B + 2\epsilon_P y z + \epsilon_R w(1-2y) \\
\frac{\partial \dot y}{\partial z} &= -\delta_P y(1-y) + \epsilon_P y^2 \\
\frac{\partial \dot y}{\partial w} &= y(1-y)(\epsilon_R - \delta_R) \\[6pt]
\frac{\partial \dot z}{\partial y} &= -zF - zy\,B + \epsilon_P z(z-1) - \epsilon_R z w \\
\frac{\partial \dot z}{\partial z} &= -yF + \delta_P y z + \epsilon_P y(2z-1) + \epsilon_R w(1-y) \\
\frac{\partial \dot z}{\partial w} &= \delta_R y z + \epsilon_R z(1-y) \\[6pt]
\frac{\partial \dot w}{\partial y} &= -wF - wy\,B + \epsilon_P z w + \epsilon_R w(1-w) \\
\frac{\partial \dot w}{\partial z} &= y w (\delta_P + \epsilon_P) \\
\frac{\partial \dot w}{\partial w} &= -yF + \delta_R y w + \epsilon_P y z - \epsilon_R (1-y)(1-2w)
\end{align}
An equilibrium point is locally asymptotically stable if all eigenvalues of the corresponding Jacobian matrix have strictly negative real parts. Conversely, the presence of at least one eigenvalue with a positive real part implies that the equilibrium is unstable. Due to such complex formulations, it is intuitive to apply numerical tools to compute the Jacobian's eigenvalues.

\subsubsection{Analytical Results - In Presence of Institutional Incentives} \label{sect:insti_result}
For a fully interior equilibrium to exist (where all four strategies coexist, $0<x,y,z,w<1$), each strategy must share an identical expected payoff that equals the population average:  $\Pi_C^\theta=\Pi_D^\theta=\Pi_{SP}^\theta=\Pi_{SR}^\theta\,(=\bar\Pi^\theta)$. The condition gives rise to the following linear system:
\begin{align}
\text{(a)} \quad & F(y,z,w) + \gamma = 0, \label{eq:interior-a} \\
\text{(b)} \quad & -\epsilon_P y + \alpha = 0, \label{eq:interior-b} \\
\text{(c)} \quad & -\epsilon_R (1-y) + \beta = 0, \label{eq:interior-c} \\
\text{(d)} \quad & x+y+z+w=1, \label{eq:interior-d}
\end{align}
which in linear-system form (in $x,y,z,w$) reads
\begin{equation}
\begin{pmatrix}
0 & B & -\delta_P & -\delta_R \\
0 & \epsilon_P & 0 & 0 \\
0 & \epsilon_R & 0 & 0 \\
1 & 1 & 1 & 1
\end{pmatrix}
\begin{pmatrix} x \\ y \\ z \\ w \end{pmatrix}
=
\begin{pmatrix} R-T-\gamma \\ \alpha \\ \epsilon_R - \beta \\ 1 \end{pmatrix}.
\end{equation}
Unlike the peer-incentive-only case, (b) and (c) no longer force $y$ to $0$ and $1$ simultaneously: they fix $y = \alpha/\epsilon_P$ from (b) and $y = 1-\beta/\epsilon_R$ from (c). These are consistent only under the \textit{solvability condition}
\begin{equation}
\label{eq:solvability}
\frac{\theta_{SP}-\theta_C}{\epsilon_P} \;=\; 1 - \frac{\theta_{SR}-\theta_C}{\epsilon_R} \qquad \Longleftrightarrow \qquad \frac{\alpha}{\epsilon_P} + \frac{\beta}{\epsilon_R} = 1,
\end{equation}
i.e.\ the two candidate values of $y$ agree only if the incentive ratios $\alpha/\epsilon_P,\beta/\epsilon_R$ sum to $1$; if \eqref{eq:solvability} fails, (b) and (c) pin $y$ to two different values and no fully interior equilibrium exists. However, solvability alone does not say where the common value $y^\star=\alpha/\epsilon_P$ lands: since $\epsilon_P>0$, $y^\star\in(0,1)$ if and only if:
\begin{equation}
\label{eq:ystar-range}
0 \;<\; \theta_{SP}-\theta_C \;<\; \epsilon_P,
\end{equation}
which we assume holds whenever an interior equilibrium is under discussion. When \eqref{eq:solvability} and \eqref{eq:ystar-range} both hold, $y=y^\star\in(0,1)$ is pinned down, and (a) together with (d) leaves two linear equations in $(x,z,w)$ — generically a one-parameter family (a line segment) rather than an isolated point, intersected with the positivity constraints $0<x,z,w<1$. This degeneracy already appears without institutional incentives; $\theta$ only controls whether the line intersects the open interior of the simplex at all.

\subsubsection*{The full equilibrium set on the simplex}
Recall the state space is the tetrahedron $\Delta = \{(x,y,z,w)\ge 0: x+y+z+w=1\}$.
Josef Hofbauer, Karl Sigmund stated a crucial property of the replicator dynamics: Every face of the tetrahedron $\Delta$ is invariant, or that a strategy eliminated at time $t_0 \ge 0$ will remain extinct for all $t \ge t_0$ \citep{hofbauer1998evolutionary}. As a result, equilibria organise naturally by how many strategies are simultaneously present: vertices (1 strategy), edges (2 strategies), faces (3 strategies), and the open interior (all 4 strategies). Each level is simply the interior-equilibrium problem of a smaller sub-game obtained by fixing the absent strategies' frequencies to $0$, so we work through them in order of increasing dimension, reusing the same payoff-equality conditions as above. To keep the notation compact write
\begin{equation}
y^\star := \frac{\alpha}{\epsilon_P}, \qquad K := (T-R+\gamma) + B y^\star,
\end{equation}

Here $y^\star$ is the common value of $y$ forced by conditions (b)/($C$) \eqref{eq:interior-b}--\eqref{eq:interior-c} once the solvability condition \eqref{eq:solvability} holds. Substituting $y=y^\star$ into condition (a) \eqref{eq:interior-a}, $F(y,z,w)+\gamma=0$, and using $F=(T-R)+By-\delta_P z-\delta_R w$, gives $\delta_P z+\delta_R w = K$: once $y$ is pinned at $y^\star$, the remaining frequencies $z$ and $w$ are no longer independent -- they must trade off against each other, weighted by the peer-punishment/reward impacts $\delta_P,\delta_R$, so that condition (a) still holds.

\paragraph{Vertices (4).} $e_C=(1,0,0,0)$, $e_D=(0,1,0,0)$, $e_{SP}=(0,0,1,0)$, $e_{SR}=(0,0,0,1)$ are always equilibria (monomorphic populations are trivially fixed).

\paragraph{Edges (6).} On each edge the two absent strategies stay at $0$ and the dynamics reduce to a single scalar payoff-difference condition.
\begin{itemize}
\item $C$--$D$ ($z=w=0$): interior point at $y = (R-T-\gamma)/B$ if this lies in $(0,1)$; otherwise the edge flow is monotone between $e_C$ and $e_D$.
\item $D$--$SP$ ($x=w=0$, $y=1-z$): interior point at $z^\star_{D,SP} = \dfrac{T-R+\gamma+B+\epsilon_P-\alpha}{B+\delta_P+\epsilon_P}$ if in $(0,1)$; otherwise monotone between $e_D,e_{SP}$.
\item $D$--$SR$ ($x=z=0$, $y=1-w$): interior point at $w^\star_{D,SR} = \dfrac{T-R+\gamma+B-\beta}{B+\delta_R-\epsilon_R}$ if in $(0,1)$; otherwise monotone between $e_D,e_{SR}$.
\item $C$--$SP$ ($y=w=0$): $\Pi_{SP}^\theta-\Pi_C^\theta=\alpha$ identically on this edge, so $\dot z = z(1-z)\alpha$. If $\alpha=0$ ($\theta_{SP}=\theta_C$) the \emph{whole edge} is a degenerate continuum of neutral equilibria; if $\alpha\neq0$ the only equilibria are the vertices $e_C,e_{SP}$ (the edge flows monotonically toward whichever vertex $\alpha$ favors).
\item $C$--$SR$ ($y=z=0$): likewise $\dot w = w(1-w)(\beta-\epsilon_R)$; the whole edge is neutral iff $\beta=\epsilon_R$, otherwise only $e_C,e_{SR}$ are equilibria.
\item $SP$--$SR$ ($x=y=0$): likewise $\dot w = w(1-w)(\beta-\alpha-\epsilon_R)$; the whole edge is neutral iff $\beta-\alpha=\epsilon_R$, otherwise only $e_{SP},e_{SR}$ are equilibria.
\end{itemize}
So $D$'s presence is what generically breaks the degeneracy on the $C$--$D$, $D$--$SP$, $D$--$SR$ edges into an isolated interior point; the three edges not touching $D$ inherit the structural fact that $\Pi_{SP}^\theta-\Pi_C^\theta$ and $\Pi_{SR}^\theta-\Pi_C^\theta$ depend only on $y$, and collapse to a constant once $y=0$.

\paragraph{Faces (4).} Setting one frequency to $0$ leaves a 2-simplex with a payoff-equality system of two equations plus normalization, generically giving an isolated point.
\begin{itemize}
\item Face $w=0$ ($C,D,SP$; $SR$ absent): unique candidate $y=y^\star,\; z = K/\delta_P,\; x=1-y^\star-z$, valid whenever these lie in $(0,1)$.
\item Face $z=0$ ($C,D,SR$; $SP$ absent): unique candidate $y=y^\star$ (by the solvability condition), $w = K/\delta_R,\; x=1-y^\star-w$.
\item Face $y=0$ ($C,SP,SR$; $D$ absent): requires $\Pi_{SP}^\theta-\Pi_C^\theta=\alpha=0$ \emph{and} $\Pi_{SR}^\theta-\Pi_C^\theta=\beta-\epsilon_R=0$ simultaneously; generically neither holds, so this face has no interior equilibrium beyond its vertices/edges. In the non-generic case $\alpha=0,\beta=\epsilon_R$ the entire face becomes a 2-D continuum of neutral equilibria.
\item Face $x=0$ ($D,SP,SR$; $C$ absent): solving $\Pi_D^\theta=\Pi_{SP}^\theta$ and $\Pi_D^\theta=\Pi_{SR}^\theta$ together with $y+z+w=1$ gives a generically unique isolated point (three linear equations in $y,z,w$; not, in general, the same point as those on faces $w=0,z=0$ above).
\end{itemize}

\paragraph{Interior (open tetrahedron).} Condition (d) fixes $x$ by normalisation once $(y,z,w)$ are known, so a fully interior equilibrium has only three free candidates. Conditions (b) and (c) then pin $y=y^\star$, and condition (a) reduces to $\delta_P z+\delta_R w=K$: a single linear equation in $(z,w)$. Hence, whenever it exists, the interior equilibrium set is a one-parameter family rather than an isolated point. Parametrising by $z$ (so that $w=(K-\delta_P z)/\delta_R$ and $x=1-y^\star-z-w$), the interior equilibrium set is exactly the open segment
\begin{equation}
\label{eq:interior-segment}
\mathcal{L} = \left\{\left(1-y^\star-z-\tfrac{K-\delta_P z}{\delta_R},\; y^\star,\; z,\; \tfrac{K-\delta_P z}{\delta_R}\right) : z \in \Big(0,\, \tfrac{K}{\delta_P}\Big),\ \ 1-y^\star-z-\tfrac{K-\delta_P z}{\delta_R} > 0\right\}.
\end{equation}

\paragraph{Feasibility of $K$.} The two constraints in \eqref{eq:interior-segment} collapse to a single condition on $K$. The first requires the interval $(0,K/\delta_P)$ to be non-degenerate, i.e.\ $K>0$; this is not automatic, since $K$ can be negative once $\gamma=\theta_D-\theta_C$ is allowed to be negative. The second requires $x>0$ somewhere on the segment. As $x(z)$ is linear in $z$, with endpoint limits $x\to 1-y^\star-K/\delta_R$ (at $z\to0$) and $x\to 1-y^\star-K/\delta_P$ (at $z\to K/\delta_P$), this holds if and only if $x>0$ at at least one endpoint, i.e.\ $K<(1-y^\star)\max(\delta_P,\delta_R)$. Combining both constraints, the interior equilibrium set is non-empty precisely when
\begin{equation}
\label{eq:K-range}
0 \;<\; K \;<\; (1-y^\star)\,\max(\delta_P,\delta_R).
\end{equation}
In particular, solvability alone does not guarantee coexistence: if either \eqref{eq:solvability} or \eqref{eq:K-range} fails, then $\mathcal L=\emptyset$ and only the vertex/edge equilibria above remain.

\paragraph{Boundary limits.} When \eqref{eq:K-range} holds, the two ends of $\mathcal L$ approach the candidate boundary points
\begin{itemize}
    \item $z\to0$ ($w\to K/\delta_R$): the face-$z=0$ point $(1-y^\star-K/\delta_R,\;y^\star,\;0,\;K/\delta_R)$.

    \item $z\to K/\delta_P$ ($w\to0$): the face-$w=0$ point $(1-y^\star-K/\delta_P,\;y^\star,\;K/\delta_P,\;0)$.
\end{itemize}
If $x>0$ at both ends, both limits lie in the relative interior of their respective faces: they are exactly the face equilibria found above, and $\mathcal L$ runs from the $(D,C,SR)$ face to the $(D,C,SP)$ face, passing through the interior. If $x\le0$ at exactly one end, $\mathcal L$ is truncated at that end: it exits the simplex through the $x=0$ (i.e.\ $(D,SP,SR)$) face, at $z=\big(K-\delta_R(1-y^\star)\big)/(\delta_P-\delta_R)$, instead of reaching the corresponding face equilibrium.

\subsubsection{Replicator Dynamics and Equilibrium Simulation Results}

\begin{figure}[H]
    \centering
    \includegraphics[width=\linewidth]{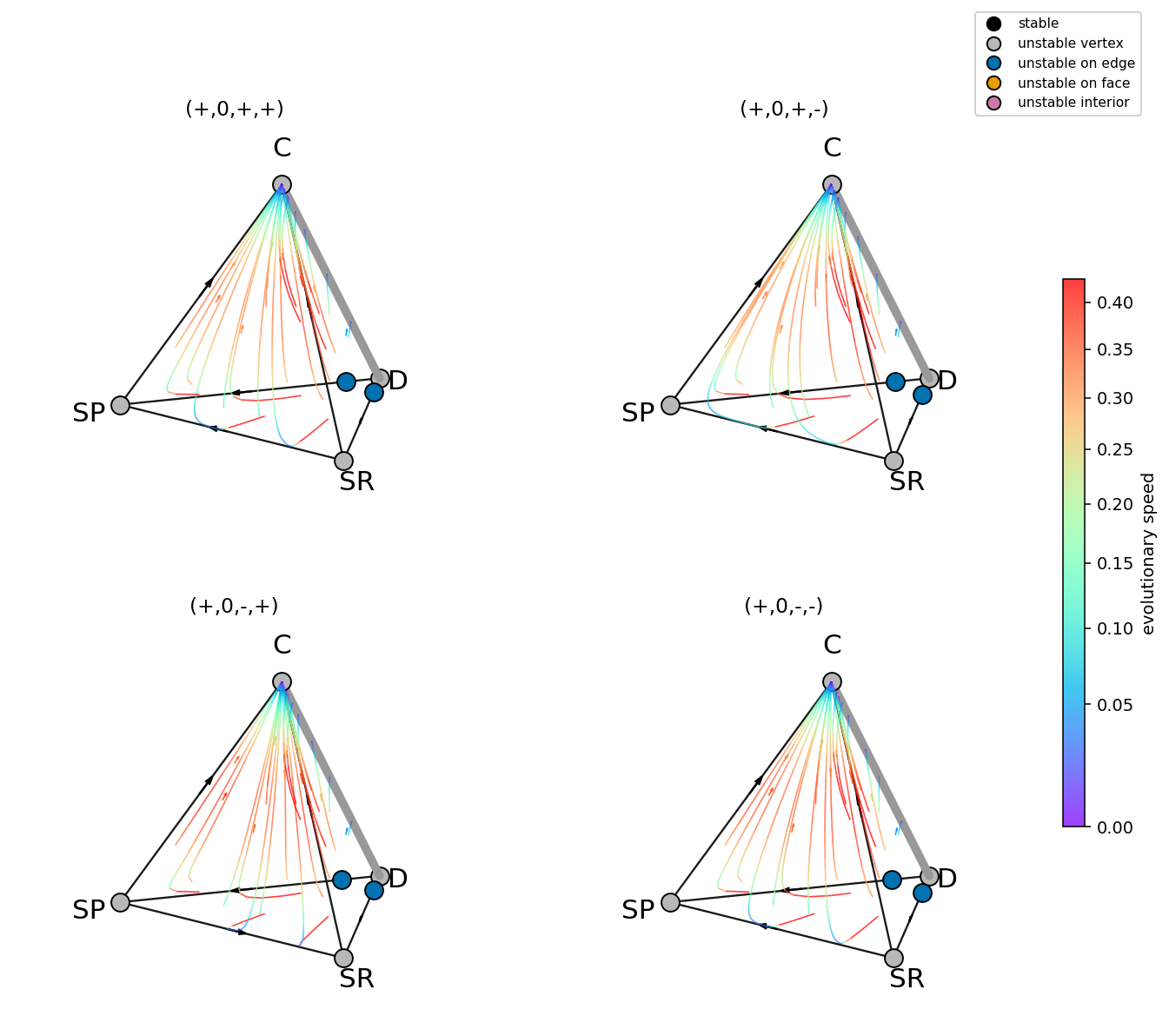}
    \caption{\textbf{Phase portraits of the four-strategy replicator dynamics on a two-dimensional simplex under the rewarding policy for cooperation ($C$).} Across all four cases, no stable equilibrium exists. Streamlines from almost all interior initial conditions converge toward $C$. The evolutionary speed is highest in regions away from the equilibrium, particularly near the center of the simplex, and gradually decreases as the trajectories approach $C$, where the dynamics eventually vanish. The trajectories along every boundary are monotonic, following a unique direction without reversals or oscillations. In addition to the four vertices, two unstable equilibrium points are located on the $D$--$SP$ and $D$--$SR$ edges, together with another unstable equilibrium in the vicinity of $D$, all having identical coordinates across the four panels. Furthermore, a continuum of equilibrium points exists on the $C$--$D$ edge, consistent with the theoretical stability analysis. The parameters are fixed at $\theta_C = 1$, $\theta_D = 0$, and $\theta_{SP}, \theta_{SR} \in \{\pm1/12\}$. Each panel's title corresponds to the values of $\theta_C$, $\theta_D$, $\theta_{SP}$, and $\theta_{SR}$, respectively.}
    \label{fig:simplex_insti_plus_zero}
\end{figure}

\begin{figure}[H]
    \centering
    \includegraphics[width=\linewidth]{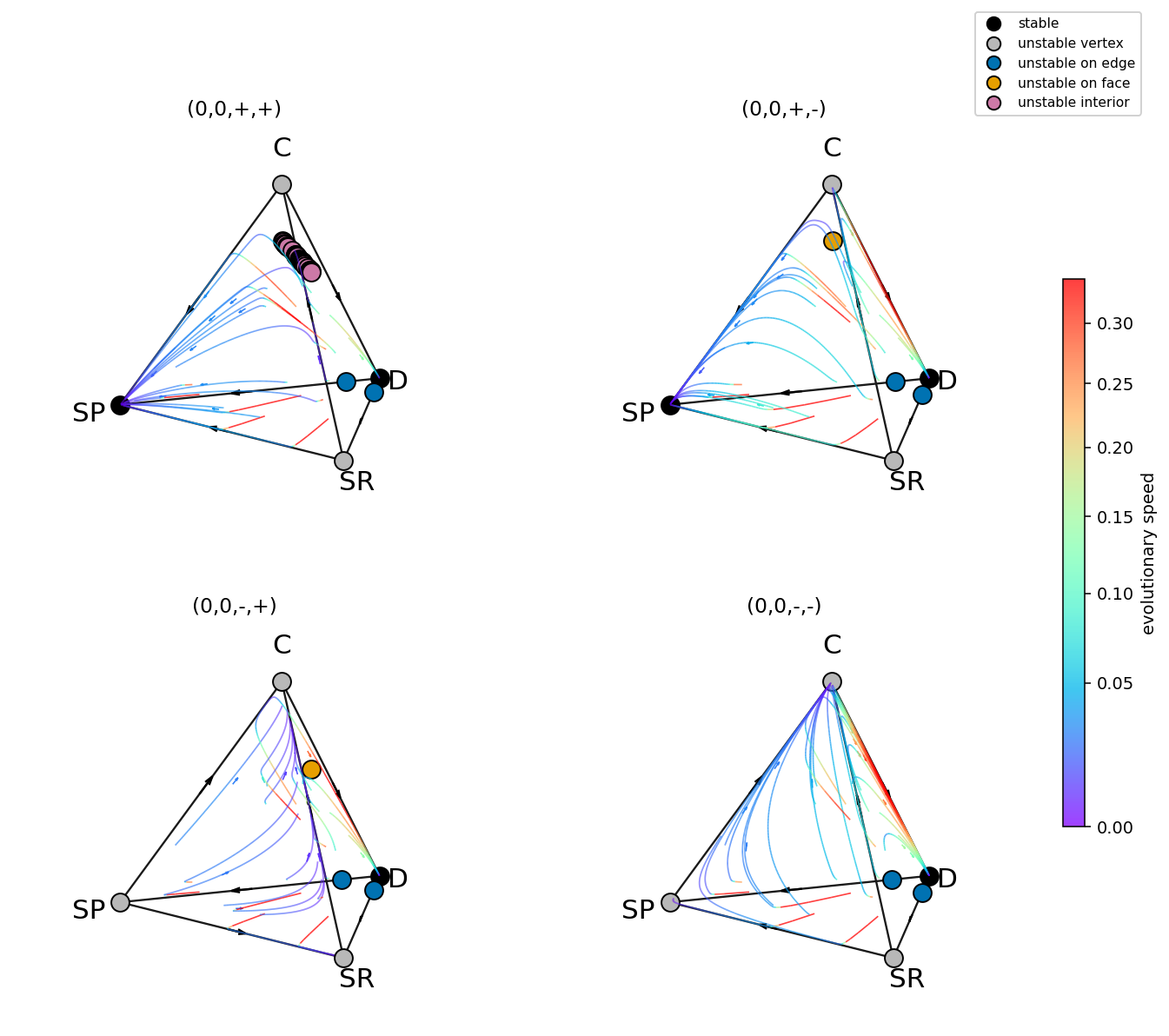}
    \caption{\textbf{Phase portraits of the four-strategy replicator dynamics on a two-dimensional simplex projection under policies that reward or punish social agents only.} Across all four cases, the stable equilibrium is the defector strategy ($D$), except for the social punisher ($SP$) in the $(0,0,+,+)$ and $(0,0,+,-)$ cases. The streamlines converge to either $D$, $SP$, or $C$, depending on the parameter configuration. The evolutionary speed gradually decreases as the trajectories approach the equilibrium and is highest in regions leading toward $D$ in all cases. In the $(0,0,-,+)$ case, the trajectories exhibit a slow rotational transient in the interior of the simplex before eventually converging to $C$. The trajectories along every boundary are monotonic, following a unique direction without reversals or oscillations. In addition to the four vertices, two unstable equilibrium points are located on the $D$--$SP$ and $D$--$SR$ edges, together with another unstable equilibrium in the vicinity of $D$. Moreover, a continuum of equilibrium points exists in the $(0,0,+,+)$ case as a line inside the tetrahedron, and an additional equilibrium point lies on the $C$--$SP$--$D$ plane in the $(0,0,+,-)$ and $(0,0,-,+)$ cases, consistent with the theoretical stability analysis. The parameters are fixed at $\theta_C = 0$, $\theta_D = 0$, and $\theta_{SP}, \theta_{SR} \in \{\pm 1/12\}$. Each panel's title corresponds to the values of $\theta_C$, $\theta_D$, $\theta_{SP}$, and $\theta_{SR}$, respectively.}
    \label{fig:simplex_insti_zero_zero}
\end{figure}
This section illustrates the population dynamics under different institutional incentive policies for $SP$ and $SR$ players, and determines whether simultaneously implementing a reward policy for $C$ players would help promote cooperation. Both Figures \ref{fig:simplex_insti_plus_zero} and \ref{fig:simplex_insti_zero_zero} displays equilibria that match the formulations in the full equilibrium set stated in section \ref{sect:insti_result}. In addition, the dynamics flow on each edge exhibits monotonic behaviour, further supporting the established analytical result on edge equilibria.

Figure \ref{fig:simplex_insti_plus_zero}, which combines a reward policy for $C$ players, yields rather advantageous results for promoting cooperation, as the dynamics show strong attraction towards vertex $C$ ($f = (1,0,0,0)^\top$). However, the magnitude of attraction plummets as the population approaches closer to vertex $C$. In addition, the numerical result shows that the equilibrium at vertex $C$ is unstable. This suggests that the policies have yet to promote cooperation optimally.

On the other hand, the dynamics in Figure \ref{fig:simplex_insti_zero_zero} evolves more slowly, indicating a lower evolutionary speed throughout the simplex. The equilibrium at vertex $D$ is consistently stable through all 4 policy regimes. However, rewarding $SP$ players obtains a rather optimistic result, as the dynamics gravitate towards a non-defector, although with lower evolutionary speed.

\subsection{Agent-based simulation results for square lattice population}
In this section, we analyse the results of the agent-based simulation on a square lattice network, through which we investigate the reciprocity of population with a co-existence of four strategies ($C$, $D$, $SP$, and $SR$) and the effect of institutional incentives and punishment on the frequency of cooperation and on social welfare.

\subsubsection{Institutional reward}


Figure \ref{fig:simulation_overview} reports the stationary cooperation level across the ($\epsilon, \delta$) plane for each reward policy. Cooperation is maximised uniformly in the low-cost, high-impact regime (small $\epsilon$ but large $\delta$) and declines as either the cost of incentivising rises or its per-unit impact falls. However, the policies differ sharply in magnitude. Specifically, rewarding cooperators alone yields cooperation levels indistinguishable from the no-institution baseline, indicating that a subsidy directed at plain cooperators does not, on its own, promote cooperation. The highest cooperation is instead achieved when the institution rewards enforcers: policies targeting $SP$, $SP$ and $SR$ jointly, or all three types attain the largest cooperative fractions, with the $SP$, $SR$ policy performing best. Punishment-oriented targeting consistently outperforms the reward-oriented counterpart ($SP$ $>$ $SR$; \{$C$, $SP$\} $>$ \{$C$, $SR$\}), and augmenting an enforcer-targeting policy with plain cooperators produces no additional gain, consistent with such budget being spent on cooperators only that do not strengthen cooperation stated above.

\begin{figure}[H]
    \centering
    \includegraphics[width=\linewidth]{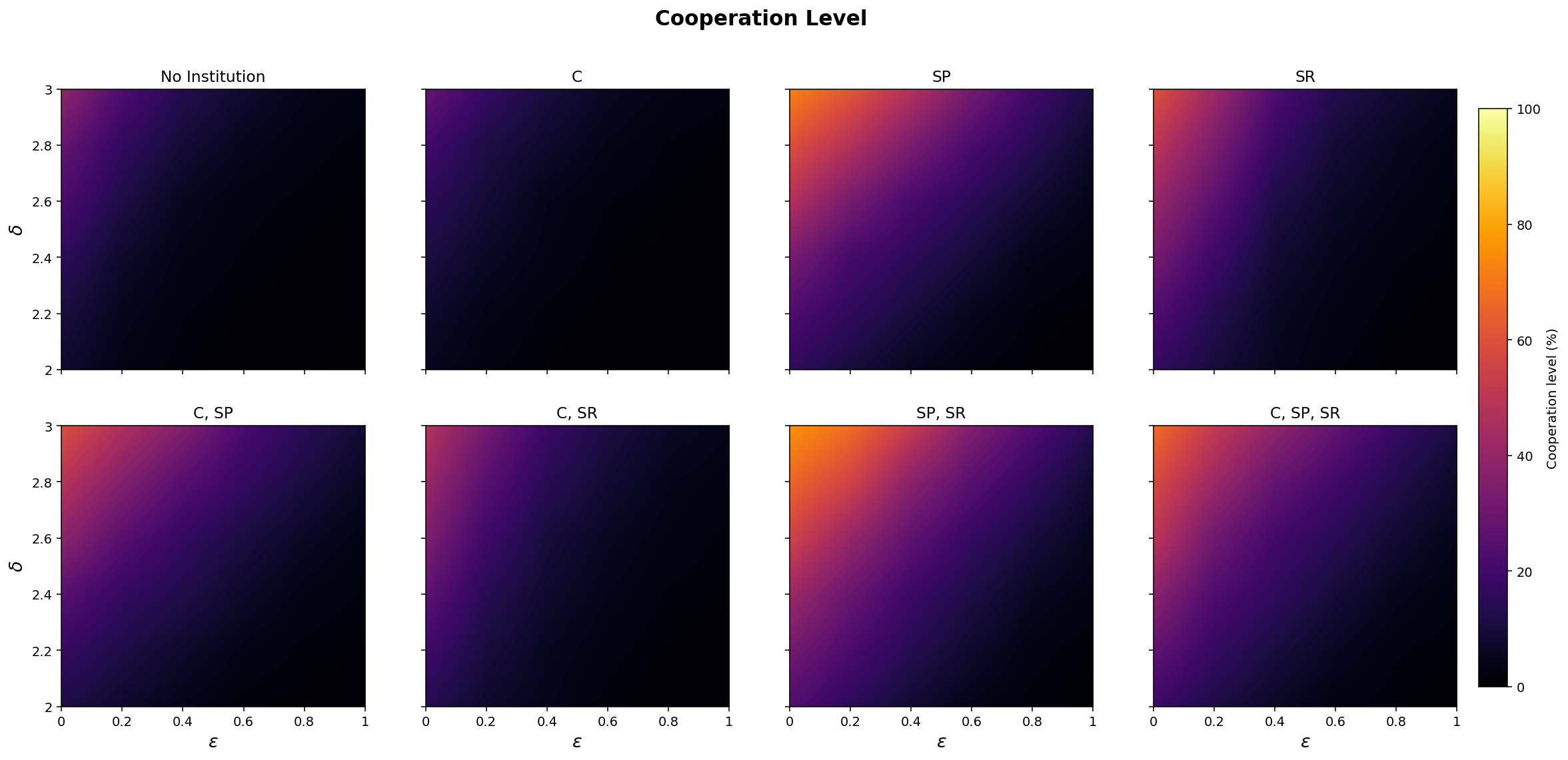}
    \caption{\textbf{Cooperation level when giving incentive on different groups of cooperators}. Each panel shows the stationary cooperation level as a function of the incentive cost $\epsilon$ (horizontal) and the impact $\delta$ (vertical), for the eight institutional targeting policies: without institution, single-target policies \{$C$\}, \{$SR$\}, \{$SP$\}, dual-target policies \{$C,SR$\}, \{$C,SP$\}, \{$SR,SP$\}, and full-target policy \{$C, SR, SP$\}. Results are averaged over 1000 last steps on an $L^2=100^2$ lattice, $\epsilon, \delta$ both in range [0, 3] (visualization is presented over a truncated range to enhance visibility).}
    \label{fig:coop_freq_incentive}
\end{figure}

\begin{figure}[H]
    \centering
    \includegraphics[width=\linewidth]{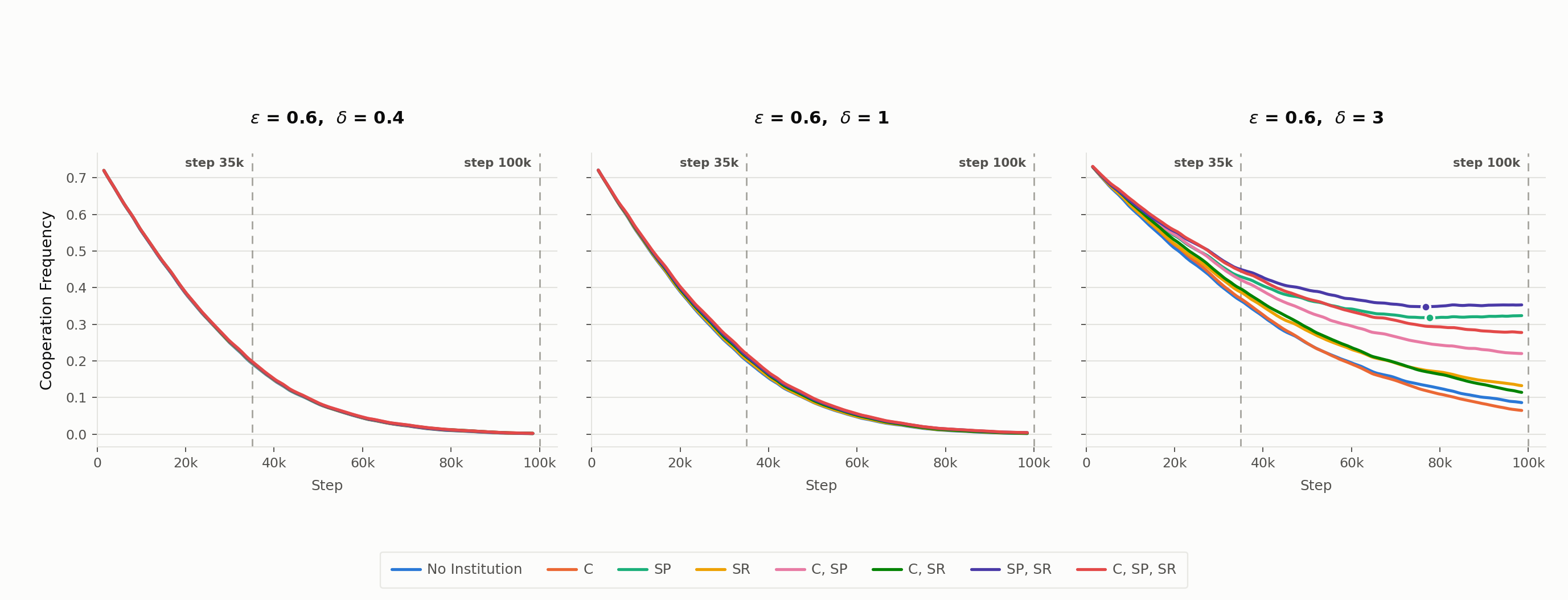}
    \caption{\textbf{Cooperation frequency in the population under different institutional regimes}. Each panel corresponds to a fixed value of $\delta$, increasing from 0.4 to 3 across the three panels, with $\epsilon$ held constant at 0.6. Within each panel, individual curves show the combined frequency of the $C$, $SP$, and $SR$ strategies under distinct institutional reward targets.}
    \label{fig:coop_freq_delta_incentive}
\end{figure}

\begin{figure}[H]
    \centering
    \includegraphics[width=\linewidth]{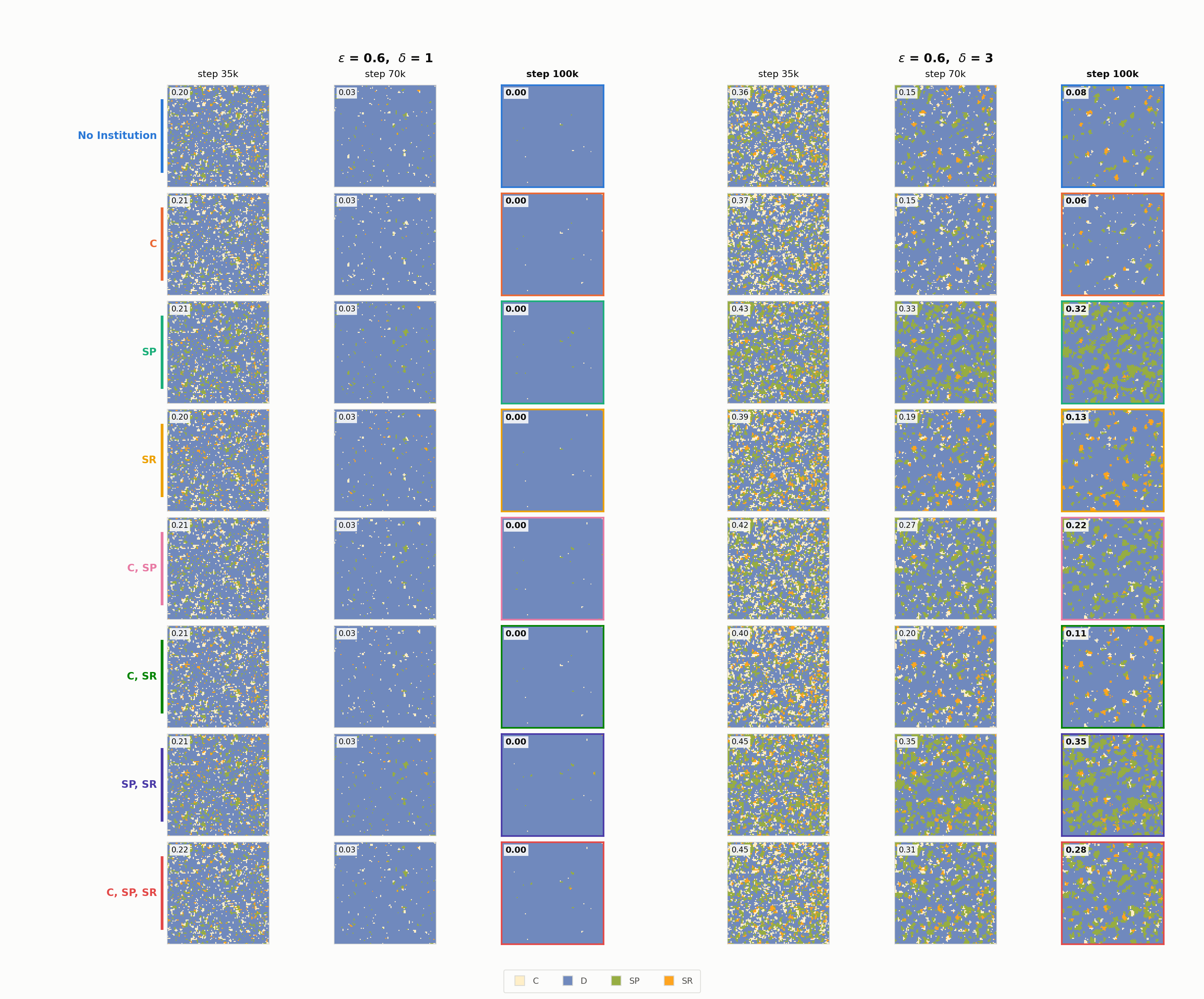}
    \caption{\textbf{Lattice representation over time under different types of incentive}. Each group of three columns shows regime with ($\epsilon, \delta$) = (0.6, 1) and (0.6, 3); within a group the columns are population states at step 35{,}000, 70{,}000 and 100{,}000. Rows are the institutional reward targets, from no institution through every combination of $C$, $SP$ and $SR$.}
    \label{fig:coop_freq_delta_incentive_lattice}
\end{figure}

With the incentive cost fixed at $\epsilon = 0.6$, the long-run cooperation level depends strongly on the incentive impact $\delta$. As shown in Figure \ref{fig:coop_freq_delta_incentive}, when $\delta = 0.4$ or $1$, all institutional targeting policies follow a similar downward trajectory, and the cooperation frequency eventually approaches zero. This pattern is consistent with the clear dominance of strategy $D$ reported in figure \ref{fig:coop_freq_delta_incentive_lattice}. In contrast, when $\delta = 3$, cooperation improves markedly under several targeting policies. In particular, $\{SP, SR\}$, $\{SP\}$, and $\{C, SP, SR\}$ achieve the highest cooperation levels. Figure~\ref{fig:coop_freq_delta_incentive_lattice} further shows that $SP$ persists and expands in the best-performing cases, all of which target $SP$. This suggests that institutionally rewarding $SP$ may help sustain cooperation when the incentive impact is sufficiently strong. Alternatively, the $C$-targeting policy does not improve cooperation relative to the no-institution baseline and even accelerate its decline. A similar pattern appears in combined targeting policies: adding $C$ consistently lowers the cooperation frequency, as indicated by $\{SP\} > \{C,SP\}$, $\{SR\} > \{C,SR\}$, and $\{SP,SR\} > \{C,SP,SR\}$. This indicates that extending institutional rewards to plain cooperators may reduce the effectiveness of rewarding incentive-providing strategies.

\begin{figure}[H]
    \centering
    \includegraphics[width=\linewidth]{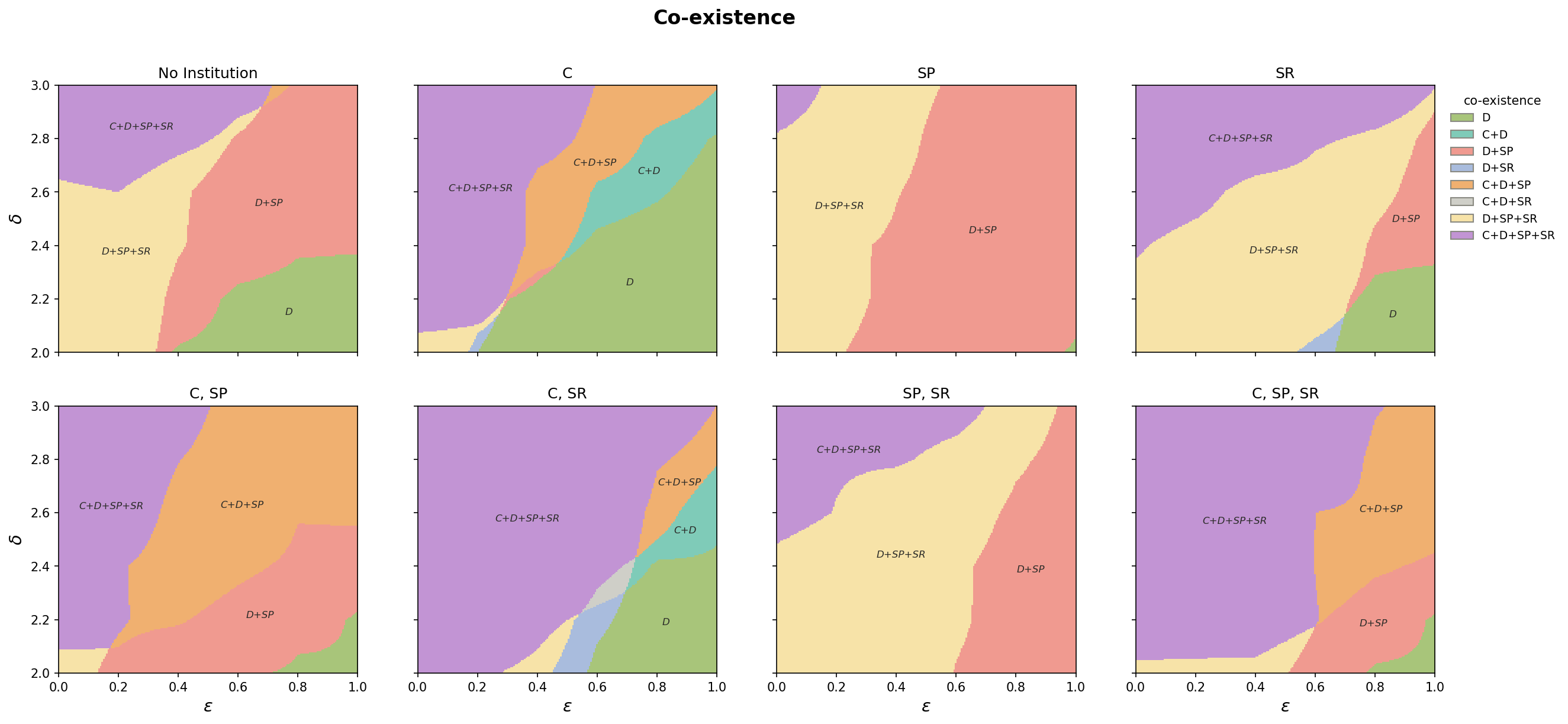}
    \caption{\textbf{Coexistence of strategies across the ($\epsilon, \delta$) parameter space under each reward policy}. Each panel maps the composition of the surviving population (the coexistence of strategies) as a function of the incentive cost $\epsilon$ (horizontal) and impact $\delta$ (vertical), for the eight institutional targeting policies: without institution, single-target policies \{$C$\}, \{$SR$\}, \{$SP$\}, dual-target policies \{$C, SR$\}, \{$C, SP$\}, \{$SR, SP$\}, and full-target policy \{$C, SR, SP$\}. Results are averaged over 1000 last steps on an $L^2=100^2$ lattice, $\epsilon, \delta$ both in range $[0, 3]$ (visualization is presented over a truncated range to enhance visibility).}
    \label{fig:co_existence_incentive}
\end{figure}

The eight panels in figure \ref{fig:co_existence_incentive} reveal distinct patterns of coexistence and strategic dominance across the ($\epsilon, \delta$) parameter space. Most notably, defectors persist in every observed outcome, no coexistence region is excluding the defector strategy. The coexistence of all four strategies is consistently concentrated in the low-cost, high-impact corner of each panel. In most panels, the opposite corner is dominated by plain defectors, reflecting conditions where incentives are costly and have limited effects. 

The largest full-coexistence region is achieved under cooperator-rewarder targeting (\{$C,SR$\}), which is matches that of the full-target policy while excluding one recipient strategy. Policies that favour plain cooperators maintain strategy C over an appreciable area, but also expand the area dominated by plain defectors. Rewarding enforcers alone (\{$SP$\}, \{$SP, SR$\}) sustains the enforcer strategies while driving C to extinction. Targeting $SP$ alone is particularly ineffective at promoting cooperation, as it leads to an extensive (\{$D, SP$\}) regime characterised by the coexistent of punishers and defectors coexist, where cooperators are no longer present. On the other hand, targeting $SR$ alone leaves the baseline structure largely unchanged.


\begin{figure}[H]
    \centering
    \includegraphics[width=0.9\linewidth]{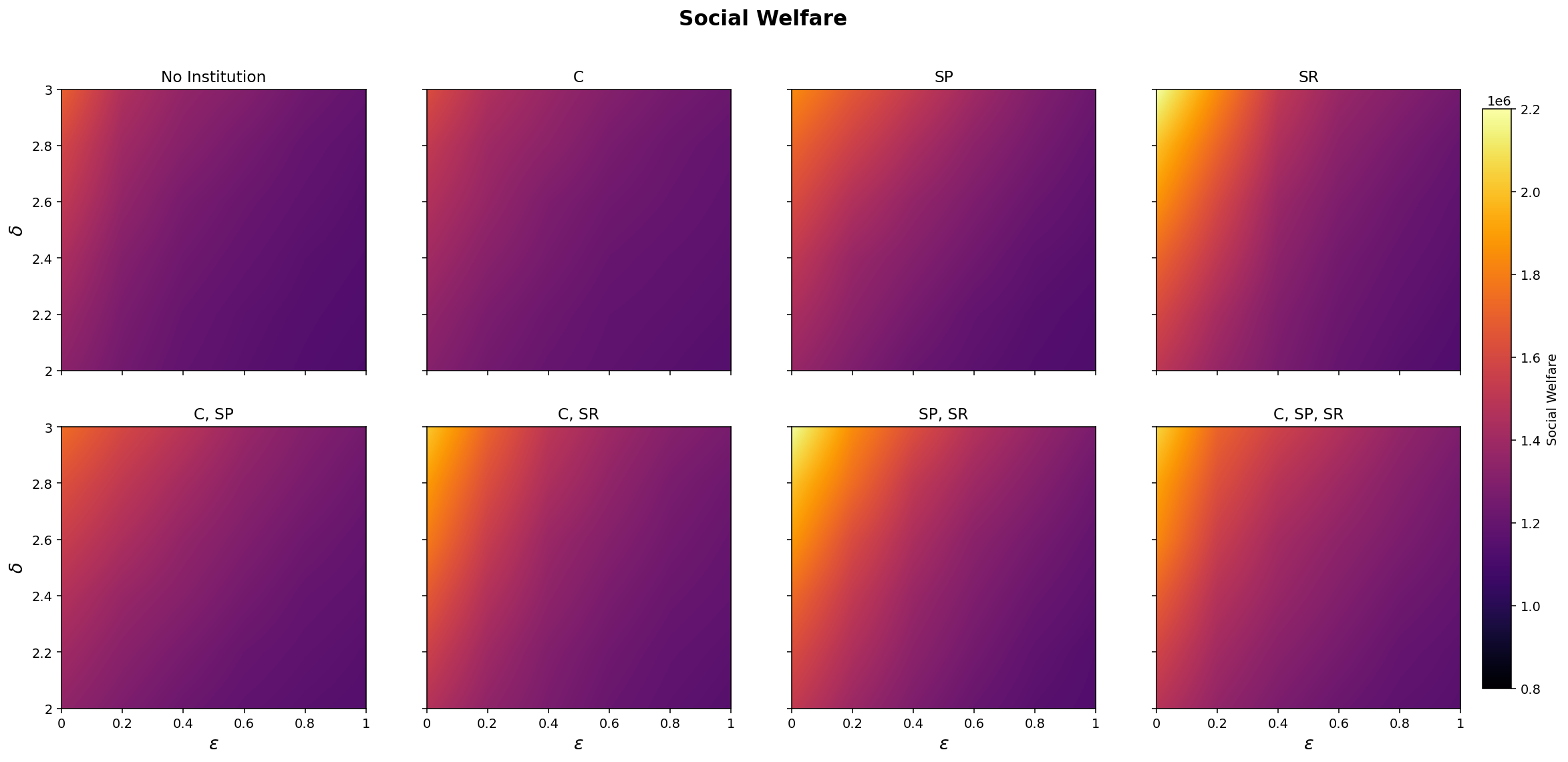}
    \caption{\textbf{Social welfare when giving incentive on different groups of cooperators}. Each panel shows the social welfare as a function of the incentive cost $\epsilon$ (horizontal) and the impact $\delta$ (vertical), for the eight institutional targeting policies: without institution, single-target policies \{C\}, \{SR\}, \{SP\}, dual-target policies \{C, SR\}, \{C, SP\}, \{SR, SP\}, and full-target policy \{C, SR, SP\}. Results are averaged over 1000 last steps on an $L^2=100^2$ lattice, $\epsilon, \delta$ both in range [0, 3] (visualization is presented over a truncated range to enhance visibility).}
    \label{fig:sw_with_diff_inst_incentive}
\end{figure}

Cost is an important factor when trying to promote social welfare, as cost increase from $\epsilon = 0$ to $\epsilon=1$ negates as much welfare as the entire range $\delta\in[2,3]$ buys back. \par
Comparing 8 scenarios inside Figure  \ref{fig:sw_with_diff_inst_incentive}, when we promote cooperators payoff, the social welfare does not benefit. This behaviour can be explained by the fact that cooperators do not contribute to generate the discipline but only the social welfare, and how defectors fully take advantage of the cooperators and gain dominance indirectly, resulting in social welfare either staying neutral or being severely affected.\par
Nevertheless, $SR$ emerges as the more favourable strategy. Although $SP$ is the most effective at maintaining the cooperation level above threshold -- as established by the cooperation-level analysis -- it does so while generating a negative-sum outcome. We cannot conclude that reward induces more cooperation than punishment; rather, reward destroys less value per unit of enforcement. Moreover, the effects of $SP$ and SR are not additive: given the presence of SR, the marginal contribution of $SP$ is approximately zero from the Figure \ref{fig:sw_with_diff_inst_incentive}

\subsubsection{Institutional punishment}

\begin{figure}[H]
    \centering
    \includegraphics[width=\linewidth]{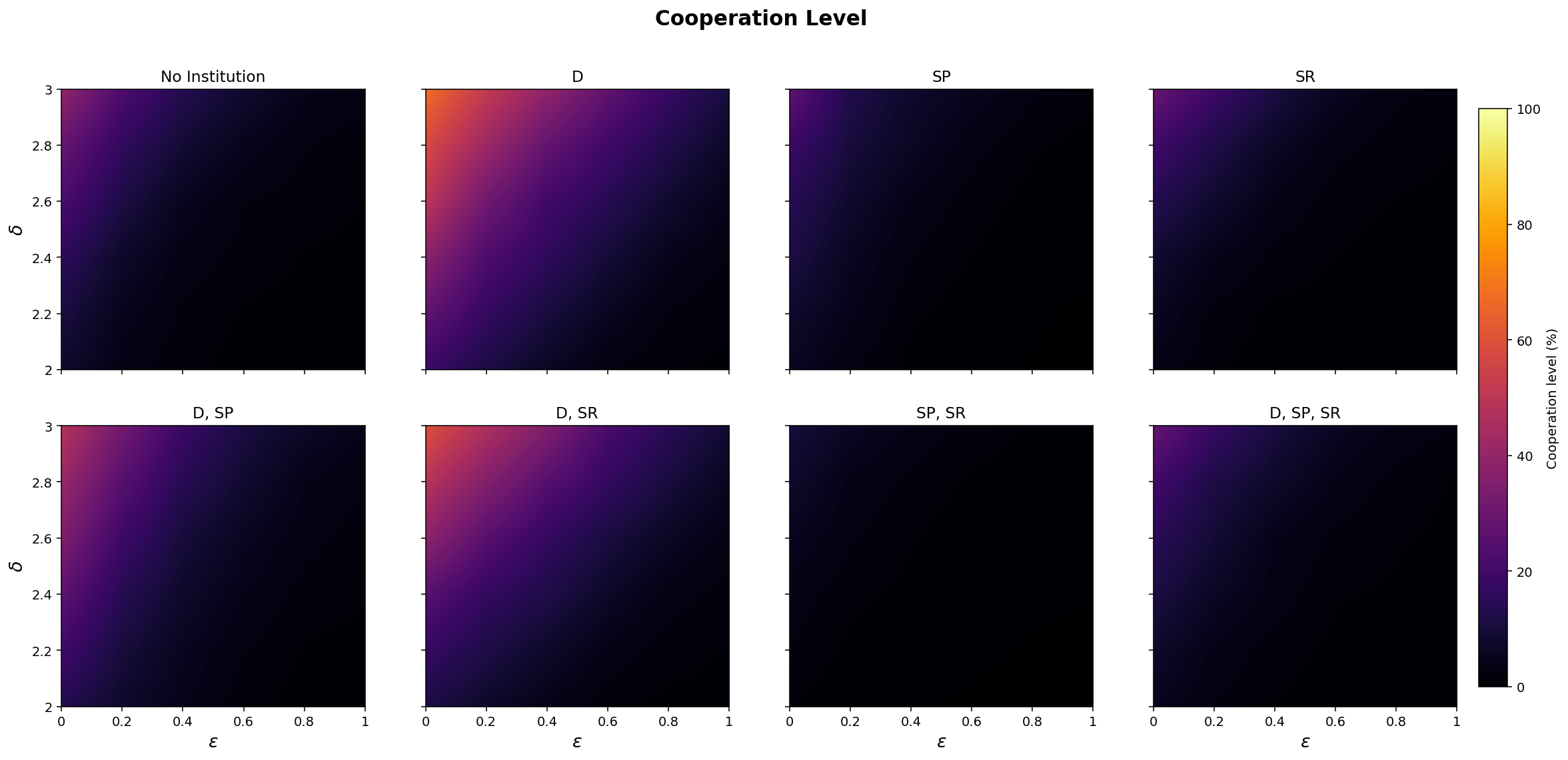}
    \caption{\textbf{Cooperation level when giving punishment on different groups of defector and social enforcers}. Each panel shows the stationary cooperation level as a function of the punishment cost $\epsilon$ (horizontal) and the impact $\delta$ (vertical), for the eight institutional targeting policies: without institution, single-target policies \{$D$\}, \{$SR$\}, \{$SP$\}, dual-target policies \{$D, SR$\}, \{$D, SP$\}, \{$SR, SP$\}, and full-target policy \{$D, SR, SP$\}. Results are averaged over 1000 last steps on an $L^2=100^2$ lattice, $\epsilon, \delta$ both in range $[0, 3]$ (visualization is presented over a truncated range to enhance visibility).}
    \label{fig:coop_freq_punish}
\end{figure}

The stationary cooperation level across the $(\epsilon, \delta)$ plane is reported in Figure~\ref{fig:coop_freq_punish} for each punishment policy. The ranking of policies is entirely reversed relative to the reward case: punishing defectors is the only effective lever. Punishing $D$ alone attains the highest cooperation across all eight panels, and every policy that includes $D$ among its targets sustains cooperation well above the no-institution baseline, whereas no policy omitting $D$ yields any improvement.

Punishing enforcers is not merely ineffective but actively destructive. Punishing $SP$ alone, as well as punishing $SP$ and $SR$ jointly, nearly drives cooperation to zero across the entire parameter space, even below the institution-free baseline. The mechanism is direct: $SP$ and $SR$ are themselves cooperators, so penalising them both removes cooperative strategies and dismantles the very peer-enforcement mechanism that had been holding defection in check. When both enforcer types are targeted simultaneously, the population collapses to a pure $C$-versus-$D$ Prisoner's Dilemma, in which defection dominates. Punishing $SR$ alone is less damaging but still leaves cooperation below baseline.

The asymmetry between the two enforcer types persists, in the same direction as under reward. Adding $SP$ to the targets alongside $D$ degrades cooperation appreciably, whereas adding $SR$ costs almost nothing. Since the cooperation lost by penalising a given type reflects how much that type contributes to sustaining it, this confirms that $SP$ plays the larger role. Punishing all three types is accordingly worse than punishing $D$ alone: budget spent penalising enforcers is both wasted and partly cancels the benefit of penalising defectors.

\begin{figure}[H]
    \centering
    \includegraphics[width=\linewidth]{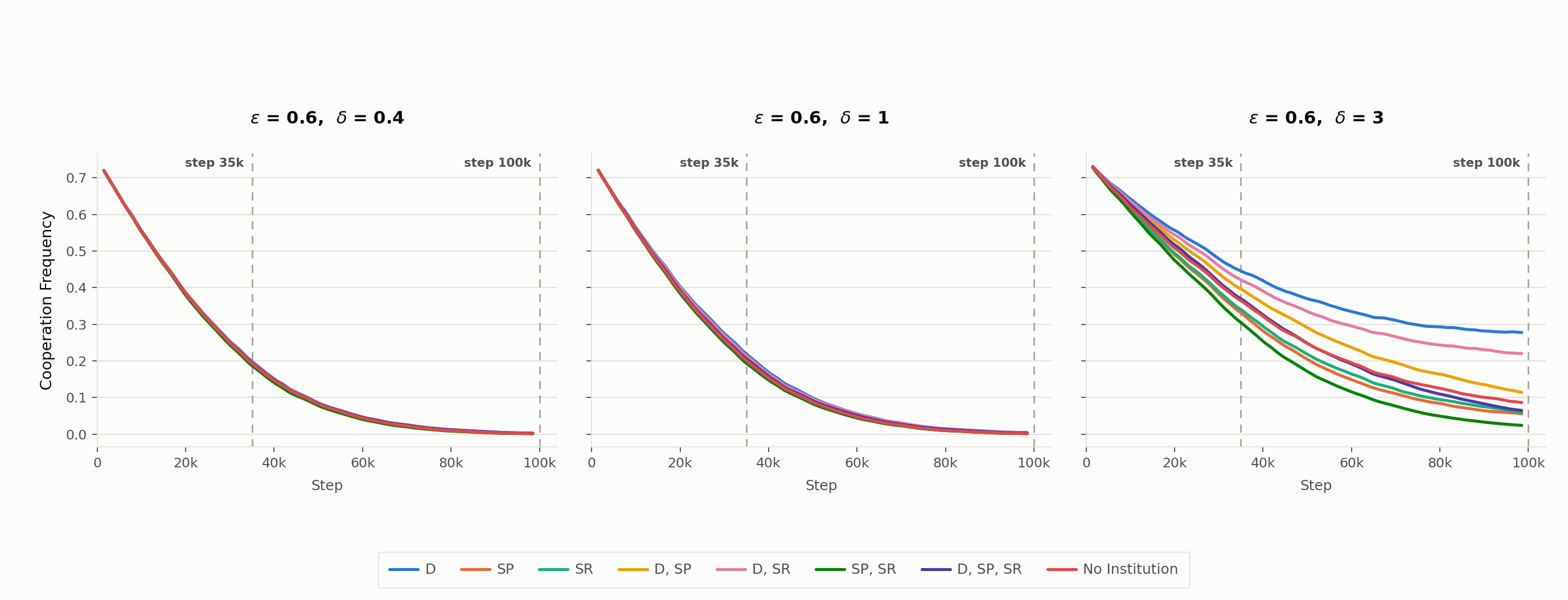}
    \caption{\textbf{Cooperation frequency in the population under different institutional punishment regimes}. Each panel corresponds to a fixed value of $\delta$, increasing from 0.4 to 3 across the three panels, with $\epsilon$ held constant at 0.6. Within each panel, individual curves show the combined frequency of the $C$, $SP$, and $SR$ strategies under distinct institutional punishment targets.}
    \label{fig:coop_freq_delta_punish}
\end{figure}

\begin{figure}[H]
    \centering
    \includegraphics[width=\linewidth]{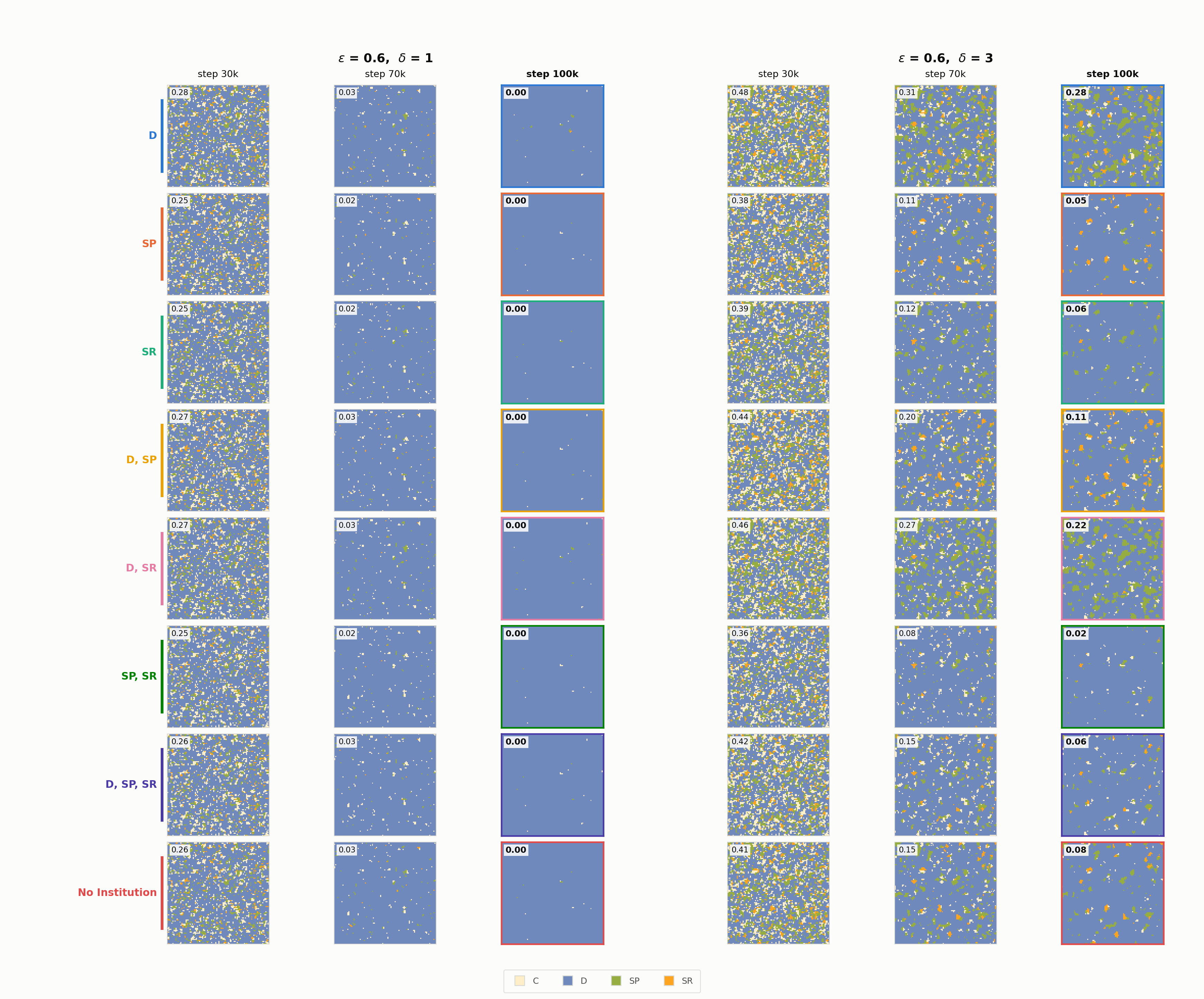}
    \caption{\textbf{Lattice representation over time under different types of punishment}. Each group of three columns shows regime with ($\epsilon, \delta$) = (0.6, 1) and (0.6, 3); within a group the columns are population states at step 35{,}000, 70{,}000 and 100{,}000. Rows are the institutional punishment targets, from no institution through every combination of $D$, $SP$ and $SR$.}
    \label{fig:coop_freq_delta_punish_lattice}
\end{figure}

Figure~\ref{fig:coop_freq_delta_punish} traces cooperation over time at $\epsilon = 0.6$ for three values of $\delta$, and shows that the policy ranking of Figure~\ref{fig:coop_freq_punish} emerges only once peer incentives are strong enough to matter. At $\delta = 0.4$ and $\delta = 1$, the eight trajectories are indistinguishable and all decay to zero, so institutional punishment leaves no imprint on the outcome. Only at $\delta = 3$ do the curves separate, and they do so gradually, remaining bunched for roughly the first 20{,}000 steps before differences accumulate. The resulting order matches the stationary picture: punishing $D$ retains the highest cooperation, followed by the policies combining $D$ with one enforcer type, while punishing enforcers alone leaves cooperation at or below the level observed without any institution. All trajectories decline monotonically from their initial value, so within this parameter range institutional punishment slows the erosion of cooperation rather than reversing it, and even the most effective policy retains under a third of the initial cooperative fraction. The underlying spatial distribution is shown in Figure~\ref{fig:coop_freq_delta_punish_lattice}.

The eight lattices are indistinguishable at every snapshot when $\delta$ is close to $\epsilon$: the mixed clusters visible at 30{,}000 dissolve steadily, and by 100{,}000 every panel has collapsed to uniform defection regardless of which types the institution targets. The choice of target set is inert in this regime, so the discussion below concerns the case where $\delta$ is well above $\epsilon$. The choice of target set is inert in this regime, so the discussion below concerns the case where $\delta$ is well above $\epsilon$.

In that case, the panels separate into three groups. Policies that punish $D$ without touching the enforcers retain the most cooperation, and the surviving structure consists of compact clusters of $SP$ and $SR$ rather than of plain cooperators, which are barely visible by 100{,}000. Policies that punish $D$ together with one enforcer type form an intermediate group in which the untargeted enforcer takes over: punishing $D$ and $SP$ leaves clusters composed almost entirely of $SR$, while punishing $D$ and $SR$ leaves clusters of $SP$, the latter retaining roughly twice the cooperation of the former. Policies that punish enforcers without touching $D$ form the lowest group, with no cluster structure remaining and a lattice essentially indistinguishable from uniform defection.

\begin{figure}[H]
    \centering
    \includegraphics[width=\linewidth]{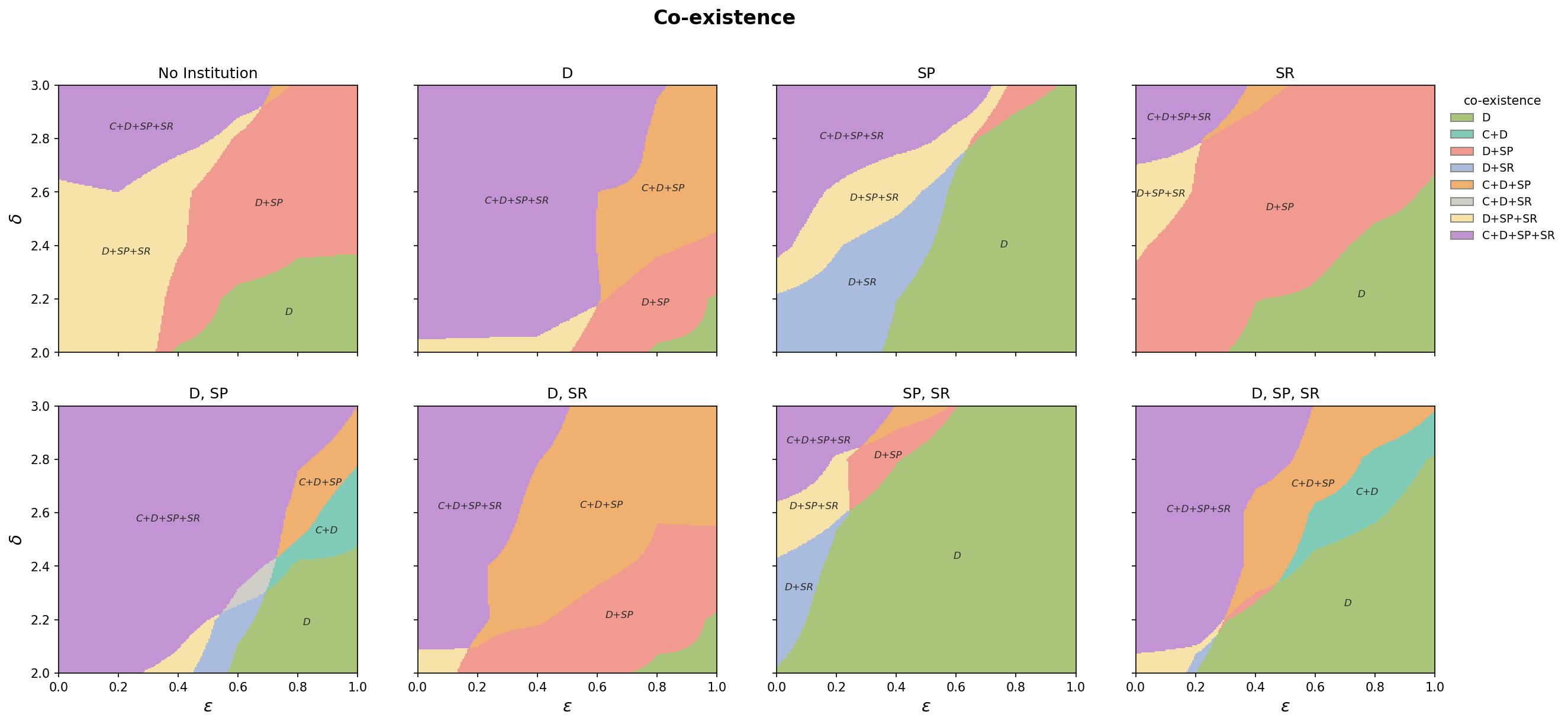}
    \caption{\textbf{Coexistence of strategies across the ($\epsilon, \delta$) parameter space under each punishment policy}. Each panel maps the composition of the surviving population (the coexistence of strategies) as a function of the punishment cost $\epsilon$ (horizontal) and impact $\delta$ (vertical), for the eight institutional targeting policies: without institution, single-target policies \{$D$\}, \{$SR$\}, \{$SP$\}, dual-target policies \{$D, SR$\}, \{$D, SP$\}, \{$SR, SP$\}, and full-target policy \{$D, SR, SP$\}. Results are averaged over 1000 last steps on an $L^2=100^2$ lattice, $\epsilon, \delta$ both in range $[0, 3]$ (visualization is presented over a truncated range to enhance visibility).}
    \label{fig:co_existence_incentive_punish}
\end{figure}

In Figure~\ref{fig:co_existence_incentive_punish}, stationary states are grouped by the set of coexisting strategies. Defectors are present in all regions of every panel, as in the reward case, suggesting that within the parameter range examined here, institutional punishment tends to reshape the composition of strategies coexisting with defection rather than remove defection itself. Only policies that include $D$ among their targets allow plain cooperators to survive. In the four panels of this kind, states containing $C$ occupy a substantial area at larger $\epsilon$, mainly $\{C, D\}$ and $\{C, D, SP\}$. In the three panels targeting enforcers alone, $C$ appears nowhere except in a small corner where all four strategies coexist. The reason is straightforward: penalising $D$ directly makes cooperation viable without any backing from peer enforcement, so $C$ persists even at large $\epsilon$, where $SP$ and $SR$ can no longer sustain themselves. Targeting enforcers has the opposite effect and collapses the population into states dominated by defection. The panel punishing $SP$ and $SR$ jointly shows the largest area occupied by $D$ alone, consistent with the vanishing cooperation levels of Figure~\ref{fig:coop_freq_punish}. The two enforcer types also appear to act as substitutes. Punishing $SR$ leaves a broad $\{D, SP\}$ region in which the surviving punishers take over the enforcement role, though not sufficiently to hold cooperation at the institution-free baseline. Punishing $SP$ shows the same pattern in reverse, leaving a narrower $\{D, SR\}$ strip.

\begin{figure}[H]
    \centering
    \includegraphics[width=0.9\linewidth]{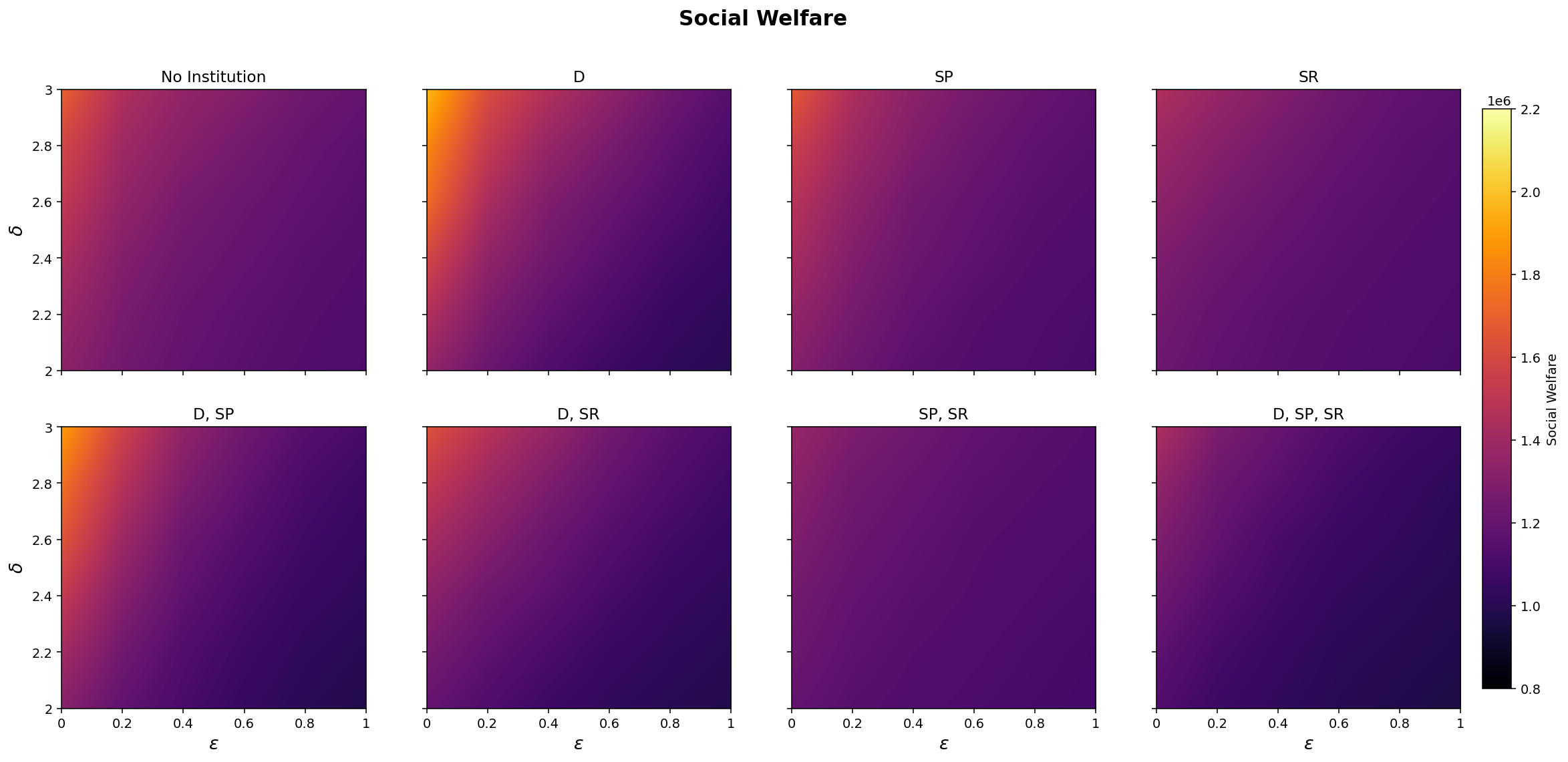}
    \caption{\textbf{Social welfare when giving punishment on different groups of defector and social enforcers}. Each panel shows the social welfare as a function of the punishment cost $\epsilon$ (horizontal) and the impact $\delta$ (vertical), for the eight institutional targeting policies: without institution, single-target policies \{$D$\}, \{$SR$\}, \{$SP$\}, dual-target policies \{$D, SR$\}, \{$D, SP$\}, \{$SR, SP$\}, and full-target policy \{$D, SR, SP$\}. Results are averaged over 1000 last steps on an $L^2=100^2$ lattice, $\epsilon, \delta$ both in range $[0, 3]$ (visualization is presented over a truncated range to enhance visibility).}
    \label{fig:sw_with_diff_inst_purnish}
\end{figure}

Figure \ref{fig:sw_with_diff_inst_purnish} shows that punish $D$ is the highest-leverage and highest-variance intervention in all experiments.This asymmetry can be explained by the fact that institutional punishment does not lower the upper hand of defectors, as $\theta=1$ is small comparing to the defector's benefit in payoff matrix. When the population shifts to a majority of $D$, the gain for social decreased as almost every interaction is also limited (they have a high probability to imitate $D$ now). \par
We also notice that punishing both enforcement channels leads to destructive damage to social welfare, with punishing $SR$ alone is as twice as harmful as $SP$. This is reasonable as we already claim that $SR$ is the key factor here since it always increase other 3 strategies balance. Pushing $SR$ will harm both the change of balance and the chance of its strategy being replicated by others. \par
There is another interesting observation: Punishing $SP$ alone does not cost overall benefit that much until we introduce punishing $D$ and $SP$. As the social in the latter interaction results in $SP$ increasing, the punishment take action and therefore the social welfare has to pay.

\section{Discussion}

Herein we have examined how peer and institutional incentives jointly shape the evolution of cooperation and social welfare in a four-strategy Prisoner’s Dilemma model with unconditional cooperators, unconditional defectors, social punishers, and social rewarders. 
Using replicator dynamics for well-mixed populations and agent-based simulations on square lattices, we showed that peer punishment is the strongest promoter of high cooperation levels, while peer reward is comparatively more beneficial for social welfare. 
For the institutional interventions, reward schemes that target peer incentive strategies, improve both cooperation and welfare, while subsidies to unconditional cooperators only have negligible and sometimes even adverse effects.
For institutional punishment, directly penalising defectors is the only consistently effective policy; punishing peer incentive strategies is detrimental for the efficiency of peer incentives, thus reducing cooperation and social welfare. 
We showed that, across both well-mixed and structured populations, defectors typically persist, and institutional interventions mainly reshape which cooperative and incentive-providing strategies coexist with them and at what welfare level.

We notice a strong relationship between the peer rewarder and the social welfare stability. This can be observed by: (1) the peer rewarder is the only strategy enhance all other strategies' gains, regardless of their intents; (2) it is a positive-sum trade off (in our setup) when social rewarder interact with any neutral or cooperate players. The strategies to increase $SR$ effectiveness and its interactions to other agents are promising to review. We also notice a clear asymmetry in how institutional reward stabilises cooperation in the well-mixed population. Rewarding $SR$ generally fails to promote cooperation, while subsiding $SP$ reverts the dynamics away from the defector equilibrium.

Our findings have several implications. First, they reinforce that designing institutions solely to maximise cooperation can be misleading: punishment-heavy regimes that sustain high cooperation may nonetheless generate low or even negative net social welfare \citep{HAN2026_social_welfare}. 
Second, by considering the co-evolution of both types of peer incentives, we clarify the distinct functional roles of social punishment and social reward: punishment is efficient at suppressing defection, while reward is better suited to preserving value once cooperation is in place. 
Third, our results suggest that institutions should prioritise penalising defectors, avoid sanctioning peer incentive strategies, and, where budgets permit, use targeted rewards for social rewarders rather than for unconditional cooperators. 
Finally, our analysis underscores the importance of network reciprocity: spatial clustering allows peer incentives and institutional policies to interact in non-trivial ways, indicating that effective real-world interventions must take underlying social structure into account \citep{duong2026beyond,szabo2007evolutionary}.

Future work could extend this framework in several directions. One direction is to let institutional incentives adapt over time and study the joint dynamics of strategies and policies. Another is to enrich the strategy set with conditional or reputation-based punishers and rewarders, or to allow the costs and impacts of incentives to evolve. It would also be valuable to test robustness on more realistic and possibly co-evolving networks, to incorporate mutation and finite-population effects more systematically, and to complement social welfare with measures of inequality. Finally, behavioural experiments could be used to validate key qualitative predictions, particularly regarding the contrasting roles of social punishment and social reward and the detrimental impact of sanctioning peer incentive strategies.

\section*{Competing Interests}
 Authors declare no competing interests.

\section*{Acknowledgements}
 T.A.H. acknowledges travel and accommodation support from the Ho Chi Minh City University of Technology (HCMUT), VNUHCM (Adjunct Professorship scheme HCMUT\text{-}VNUHCM). L.H.T acknowledges the Ho Chi Minh City University of Technology (HCMUT), VNUHCM for supporting this study. 
 TAH and Z.S. are supported by EPSRC (grant EP/Y00857X/1). MHD was supported by EPSRC grant EP/Y008561/1.



\bibliographystyle{apalike}

\bibliography{mybibfile}

\end{document}